\documentclass{article} 
\usepackage{iclr2025_conference,times}

\usepackage{amsmath,amsfonts,bm}

\def\eqref#1{equation~\ref{#1}}

\def\1{\bm{1}}

\DeclareMathAlphabet{\mathsfit}{\encodingdefault}{\sfdefault}{m}{sl}
\SetMathAlphabet{\mathsfit}{bold}{\encodingdefault}{\sfdefault}{bx}{n}

\usepackage[T1]{fontenc}
\usepackage{hyperref}
\usepackage{url}
\usepackage{array}
\usepackage{graphicx}
\usepackage{xcolor}
\usepackage{listings}
\usepackage{booktabs}
\usepackage{multirow}
\usepackage{amssymb} 
\usepackage{pifont}  
\usepackage{fontawesome5} 

\newcommand{\blfootnote}[1]{%
  \begingroup
  \renewcommand{\thefootnote}{}%
  \footnotetext{#1}%
  \endgroup
  \addtocounter{footnote}{0}%
}

\definecolor{promptbg}{HTML}{F7F8FA}
\definecolor{promptframe}{HTML}{D0D5DD}
\definecolor{promptkey}{HTML}{2C7A7B}
\definecolor{passgreen}{RGB}{46,139,87}
\definecolor{brainpurple}{HTML}{754ED0}
\lstdefinestyle{promptbox}{
  basicstyle=\ttfamily\small,
  backgroundcolor=\color{promptbg},
  frame=single,
  rulecolor=\color{promptframe},
  framerule=0.4pt,
  framesep=6pt,
  xleftmargin=12pt,
  xrightmargin=4pt,
  aboveskip=8pt,
  belowskip=10pt,
  breaklines=true,
  breakatwhitespace=true,
  columns=fullflexible,
  keepspaces=true,
  showstringspaces=false,
  upquote=true,
  literate=
    {\#}{{{\color{promptkey}\#}}}1
}
\newcommand{\bpwordmark}{\textcolor{brainpurple}{Brain}Pilot}
\newcommand{\bpbenchwordmark}{\textcolor{brainpurple}{Brain}PilotBench}

\newcommand{\dkoff}{\mbox{\faDatabase\,$-$}}
\newcommand{\dkon}{\mbox{\faDatabase\,+}}

\title{\raisebox{-0.28\height}{\includegraphics[height=1.35em]{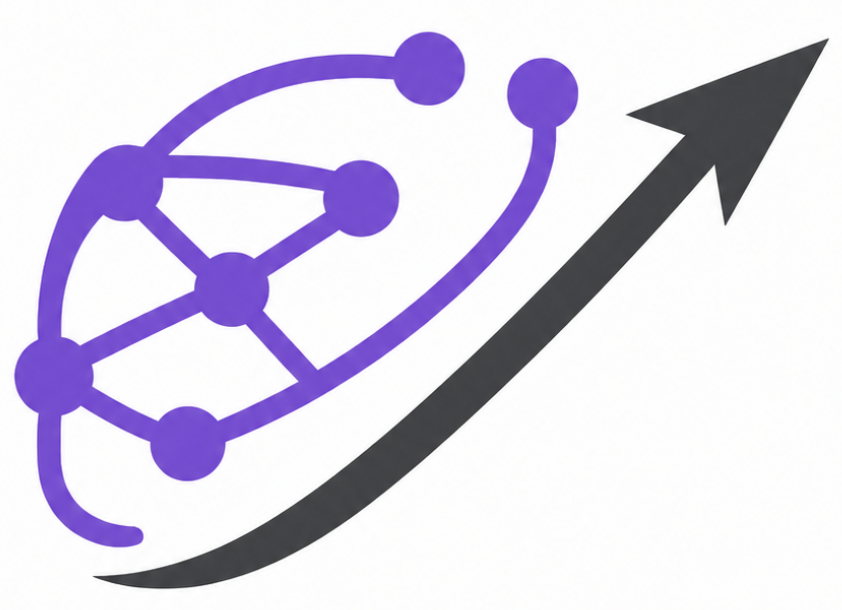}}\hspace{0.3em}BrainPilot: Automating Brain Discovery with Agentic Research}

\author{%
\normalfont
\parbox{\dimexpr\textwidth-2\tabcolsep\relax}{\raggedright
Haoxuan Li\textsuperscript{1,2,*},
Tianci Gao\textsuperscript{2,3,*},
Jianhe Li\textsuperscript{4,*},
Yang Fan\textsuperscript{5,*},
Runze Shi\textsuperscript{1,2},
Weiran Wang\textsuperscript{6},
Tianxiang Zhao\textsuperscript{2,7},
Zezhao Wu\textsuperscript{8},
Xiaoyang Jiang\textsuperscript{9},
Qihui Zhang\textsuperscript{10},
Jia Li\textsuperscript{1}, 
Xiao Xiao\textsuperscript{6},\\
Kai Du\textsuperscript{11},
Xiaoxuan Jia\textsuperscript{8},
Chao Xie\textsuperscript{11},
Lu Mi\textsuperscript{1,2,\,\faEnvelope[regular]}
\\[0.5em]
{\normalsize
\textsuperscript{1}College of AI, Tsinghua University \\
\textsuperscript{2}Shanghai Qizhi Institute \\
\textsuperscript{3}Business School, Renmin University of China \\
\textsuperscript{4}School of Physics, Beihang University \\
\textsuperscript{5}School of Information and Software Engineering, University of Electronic Science and Technology of China \\
\textsuperscript{6}Behavioral and Cognitive Neuroscience Center, Institute of Science and Technology for Brain-Inspired Intelligence, Fudan University \\
\textsuperscript{7}College of Engineering, Georgia Institute of Technology \\
\textsuperscript{8}School of Life Sciences \& IDG/McGovern Institute for Brain Research, Tsinghua University \\
\textsuperscript{9}School of Computing and Artificial Intelligence, Southwest Jiaotong University \\
\textsuperscript{10}Weixian College, Tsinghua University \\
\textsuperscript{11}Department of Psychological and Cognitive Sciences, Tsinghua University
}}%
}

\iclrfinalcopy
\begin{document}

\maketitle
\blfootnote{\textsuperscript{*}These authors contributed equally.}
\blfootnote{\faEnvelope[regular] \url{thu\_neuroai@mail.tsinghua.edu.cn}, \includegraphics[height=0.9em]{figs/icon.png}\hspace{0.1em} \url{https://brainpilot.chat}}
\setlength{\headheight}{16pt}
\addtolength{\topmargin}{-4pt}
\lhead{\includegraphics[height=11pt]{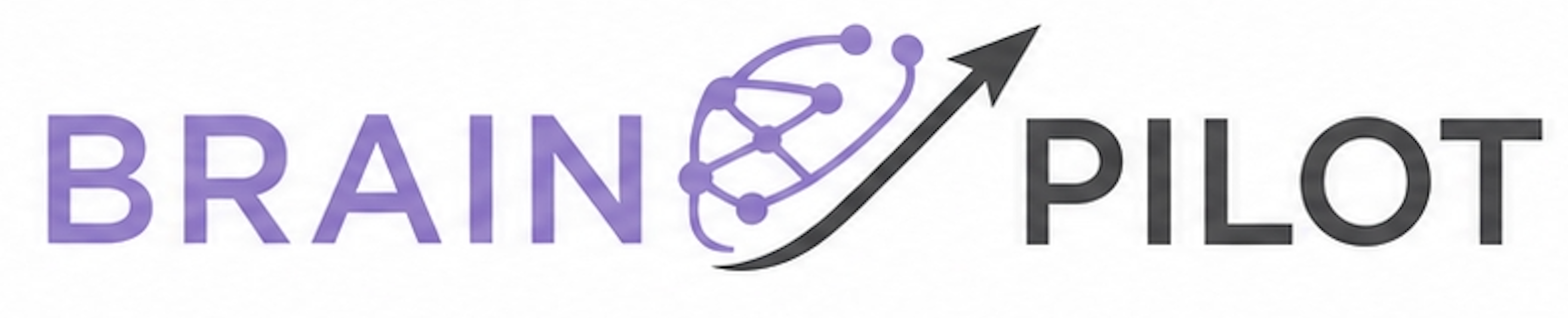}}
\rhead{\small 2026-07-16}
\vspace{-3mm}
\begin{abstract}
Understanding the brain increasingly depends on integrating evidence across
scales, modalities, and disciplines. Addressing a single research question
therefore requires a coordinated sequence of operations, from surveying prior
work to executing analyses and interpreting results in light of domain
knowledge. AI
agents promise to accelerate this process, but current agents lack domain
expertise in brain science, may fabricate claims, drift during multi-step
reasoning, and offer few defined points for expert intervention. These failures
are especially costly in brain science, where conclusions feed into downstream
scientific claims and depend on laboratory-specific expertise and careful human
judgment. We present \textbf{BrainPilot}\footnote{\url{https://github.com/NeuroAIHub/BrainPilot}},
a \textbf{fully open-source} multi-agent system that accelerates brain science research
with traceable logs and agent-verified results. A principal investigator (PI)
agent coordinates specialist agents grounded in curated domain knowledge:
a unified brain science knowledge base containing 7{,}233 indexed items and a
skill library of 72 reusable methodology units across seven research domains.
Every major step is recorded in the Graph of Trace,
an auditable record that links subgoals, tool use, evidence, and claims and
allows researchers to follow and inspect the workflow. An Auditor agent further
integrates fabrication checking into the workflow. For evaluation, we run three
brain science tasks from Agents' Last Exam, introduce our own benchmark,
\textbf{BrainPilotBench-v0}\footnote{\url{https://github.com/NeuroAIHub/BrainPilotBench}},
and present additional end-to-end case studies. Across these evaluations,
BrainPilot with an open-source backbone model attains performance comparable to state-of-the-art agent framework with less costs.
\end{abstract}

\section{Introduction}
\label{sec:intro}


For a century of neuroscience studies, scientists have pursued an enduring mission of understanding
the nature of intelligence. Rapid advances in artificial intelligence (AI) have
renewed attention to what intelligence is and how it arises. Yet progress in brain science increasingly depends on integrating evidence
across scales, modalities, and disciplines
\citep{akil2011mining,zeng2017celltypes,paninski2018neuraldatascience}. Modern brain
research is consequently a data- and coordination-intensive enterprise,
drawing on rapidly expanding literatures, heterogeneous datasets, specialized
analytical tools, and community-scale initiatives
\citep{poldrack2014bigdata,grillner2016initiatives}.
Brain science research requires coordinating literature retrieval, experimental design, data analysis, and scientific interpretation across multiple modalities\citep{biccn2021multimodal}.  These operations are time-consuming, difficult to reproduce, and
sensitive to analytical choices
\citep{fregnac2017industrialization,urai2022largescale}.

\begin{figure}[t]
\centering
\includegraphics[width=\linewidth]{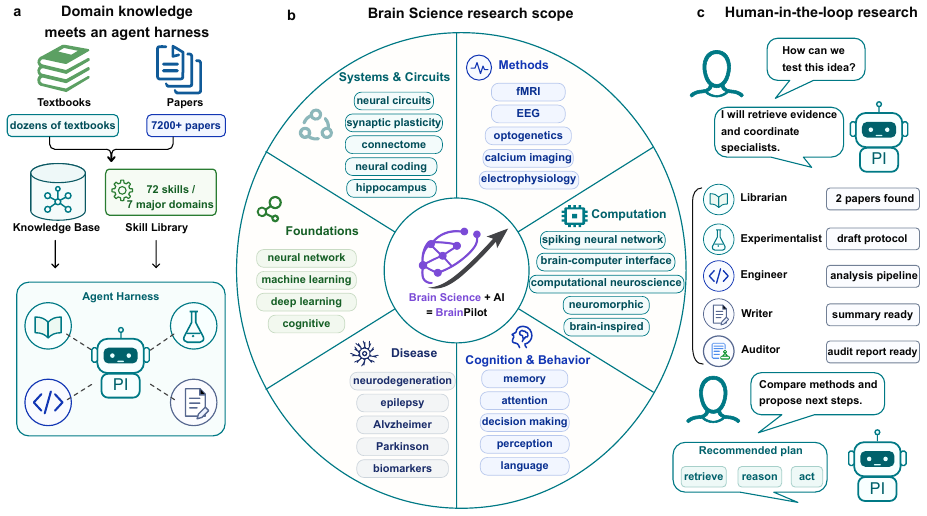}
\caption{\textbf{Overview of BrainPilot.} \textbf{(a)} A curated knowledge base
and skill library, built from neuroscience textbooks and papers, feed a harness
in which a principal investigator (PI) agent coordinates specialist agents.
\textbf{(b)} The targeted breadth of brain science research, from systems and
circuits to cognition, disease, and foundational machine learning. \textbf{(c)}
The researcher interacts with the PI agent, which delegates tasks to specialist
agents while keeping the researcher in the loop.}
\label{fig:overview}
\end{figure}

\paragraph{The lack of advances in brain science.} The emergence of AI agents brings timely technical capability to meet scientific need, where brain science increasingly requires systems that can organize
and accelerate complex research workflows while preserving the expert judgment
on which the field depends. Scientific agents are becoming practical tools for
accelerating discovery across multiple domains. In chemistry, autonomous agents
have been used to design, plan, and perform experiments by combining language
models with web search, documentation retrieval, code execution, and laboratory
automation
\citep{boiko2023coscientist, bran2024chemcrow}. In broader computational
science, tool-oriented agents have been developed to select and execute
specialized scientific tools across domains \citep{huang2025biomni}. In brain science, language models
trained or evaluated on the literature have shown promise in predicting
experimental outcomes, suggesting that large-scale models can help synthesize
prior findings in knowledge-intensive fields \citep{luo2025brainbench}, but
this addresses only one part of brain science within the broader multi-step research process. 

\paragraph{The importance of scientific validity.} The above works establish the feasibility of agent-assisted science, yet advances remains immature from another important perspective:
agents can fabricate claims, drift across multi-step reasoning, and offer few clearly defined points for expert intervention
\citep{liu2023trustworthy, lin2025agenthalluc}. These weaknesses are amplified
in brain science, where laboratory-specific expertise, careful controls, and
iterative experts' judgment are essential. 
For AI-assisted acceleration to be useful in brain science, the research
process must remain trustworthy, and human expertise must remain in the loop at
points where it can shape decisions. Large-scale evidence shows that human--AI
combinations
outperform either alone only when the collaboration is carefully designed
\citep{vaccaro2024humanai}. This finding motivates interfaces through which
experts can inspect and guide the process while evaluating the final answer
\citep{shao2025sciscigpt}. Scientific conclusions depend on the
evidence, assumptions, operations, and judgments that lead to them, as well as
on the final answer. 
A system that searches the literature, selects analyses,
invokes tools, and generates interpretations should therefore reveal how
intermediate results are produced and how they support later claims. 
This requirement is especially important in brain science, where heterogeneous
evidence often admits multiple interpretations and expert judgment is needed
to evaluate methodological choices, experimental controls, and the plausibility
of conclusions~\citep{vaccaro2024humanai}.
Much of the relevant
expertise is also laboratory-specific and poorly documented: procedures,
defaults, and rules of thumb vary across laboratories and rarely enter the
literature on which large language models are trained, making a general-purpose model poor stand-in for the field.

These demands make an AI system for brain science research more than a general-purpose system applied to a new domain and directly shape its design: curated knowledge and reusable skills provide domain grounding, PI-led specialist coordination maintains coherence across multi-step workflows, and dedicated modules expose evidence and claims for expert review while reconciling evidence across scales and modalities, preserving methodological constraints across long workflows, and supporting decisions whose validity depends on domain expertise and experimental context.

Here we present \textbf{BrainPilot}, a human-in-the-loop multi-agent system for
accelerating brain science research by coordinating domain knowledge,
analytical skills, and specialist agents within a traceable workflow in which
intermediate actions, evidence, and claims remain available for expert
inspection and evaluation. (1) BrainPilot organizes research tasks through a
principal investigator (PI) agent and specialist agents for literature search,
experimental and analytical planning, execution, interpretation, and audit.
(2) The agents are grounded in curated domain knowledge assembled locally rather
than relying on what the underlying language model happens to remember: a
unified brain science knowledge base containing 7{,}233 indexed items and a
skill library of 72 reusable methodology units across seven research domains.
(3) Their actions are recorded as a structured
workflow trace that links subgoals, tool use, evidence, and generated claims.
This design allows researchers to follow a task, examine the basis of
intermediate outputs, and apply domain judgment throughout the process.
(4) We evaluate BrainPilot on the three brain science related studies from Agents' Last Exam~\citep{ale2026}, as well as our newly proposed benchmark \textbf{BrainPilotBench-v0}, with a preliminary version including four scientific tasks that evaluate agents in brain science studies (Section~\ref{sec:benchmark}), we also conducted five case studies to evaluate BrainPilot in the real-world brain science research workflows.

\section{Related Work}
\label{sec:related}

\subsection{AI for science}

Large language model have been applied across scientific domains to
automate parts of the research process
\citep{xi2023rise, guo2024multiagent, wei2025agenticscience, zhang2025llmscimethod}.
In chemistry, autonomous agents combine language models with curated tools,
documentation retrieval, code execution, and robotic platforms to plan and
carry out experiments
\citep{boiko2023coscientist, bran2024chemcrow}, and autonomous laboratories
couple such planning with robotic synthesis to run experiments end to end
\citep{szymanski2023alab}. In biomedicine, general-purpose agents work with
large curated collections of tools, software, and databases to perform tasks
such as gene prioritization and drug repurposing \citep{huang2025biomni}, while
multi-agent systems generate hypotheses, design experiments, and analyze
results across the discovery cycle
\citep{ghareeb2026robin, gottweis2026coscientist}. More broadly, agentic systems
have moved toward automating the research process itself, from
idea to manuscript \citep{lu2024aiscientist, lu2026endtoend}. Foundation models
for biomolecules, including models that predict protein structures at
near-experimental accuracy, show that learned models can solve long-standing
scientific problems \citep{jumper2021alphafold}. These systems demonstrate that
agents and models can perform substantial parts of scientific workflows. They
are designed primarily to maximize task performance and autonomy, recording
their intermediate steps as logs.

\subsection{AI agent for brain science}

A growing number of systems bring agentic AI to brain science. Some automate
neuroethology and animal-behavior analysis end to end, from pose
estimation through statistical profiling to report generation
\citep{chen2026ethoclaw}. Others operate directly on raw multimodal neuroimaging
data and orchestrate reproducible analysis pipelines
\citep{wang2026neuroclaw}. Multi-agent frameworks have also been built for
autonomous brain-signal understanding, decomposing brain--computer-interface
workflows into specialized sub-agents under a central supervisor
\citep{zhou2026brainagent}, and agent-driven systems have been applied to the
science of mind, conjecturing and testing computational models of human
behavior \citep{prystawski2026autopsych}. These systems show that specialized
agents can carry brain science analyses through many steps with limited manual
effort. They focus on particular analysis pipelines, while the multi-step
process by which a result is reached remains largely internal to the system.
This matters because agentic systems remain prone to hallucination, instruction
drift, and silent error, which the literature on trustworthy and reliable
agents identifies as central obstacles to dependable deployment
\citep{liu2023trustworthy, yu2025trustagents, lin2025agenthalluc}. 

\subsection{Benchmarks for scientific agents}

Evaluating such systems has motivated several types of benchmarks. Knowledge
benchmarks test factual and reasoning ability across many
subjects, including biology and medicine \citep{hendrycks2021mmlu}. Task
benchmarks evaluate agents on data-driven scientific coding and discovery,
scoring each task by executing the generated programs
\citep{chen2024scienceagentbench, majumder2024discoverybench, li2025autosdt}.
Recent evaluations extend from isolated tasks to longer research workflows. A
recent review distinguishes discovery, execution and verification, synthesis,
and provenance-oriented instruments~\citep{tie2026autoresearch}. AIRS-Bench
covers idea generation, experimental analysis, and iterative refinement across
20 machine-learning research tasks, while FIRE-Bench evaluates whether agents
can rediscover published scientific insights through autonomous experimentation
\citep{lupidi2026airsbench, wang2026firebench}. ResearchClawBench evaluates
end-to-end research on 40 paper-grounded tasks across 10 scientific domains
using expert-curated multimodal rubrics~\citep{xu2026researchclawbench}, and
NatureBench packages 90 tasks from Nature-family papers into containerized
environments with hidden test sets and automated evaluators
\citep{wang2026naturebench}.
For brain science, BrainBench evaluates whether models can predict the
outcomes of neuroscience experiments from the literature
\citep{luo2025brainbench}. Together, these resources move evaluation toward
paper-grounded, multi-step research, but remain general, machine-learning
centric, or cross-domain.

\section{System}
\label{sec:system}

\begin{figure}[t]
\centering
\includegraphics[width=\linewidth]{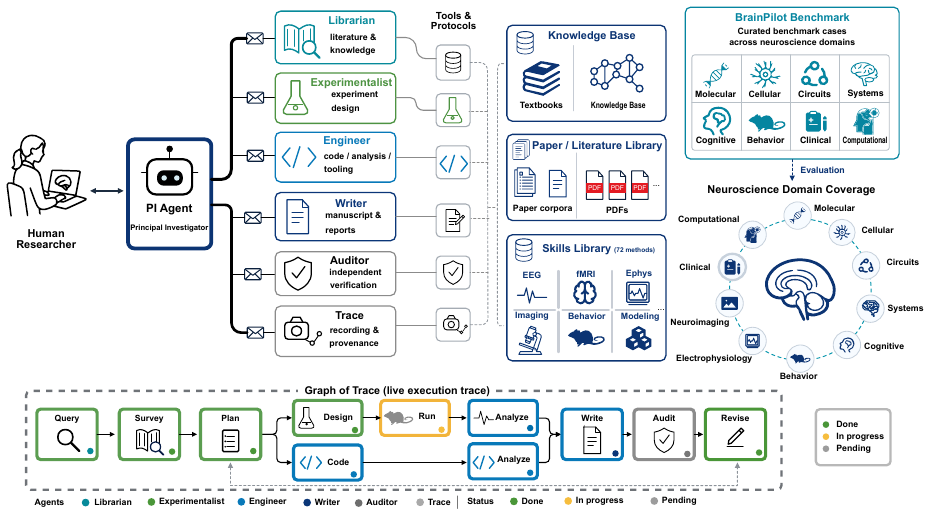}
\caption{\textbf{BrainPilot system architecture.} The researcher interacts with
the principal investigator (PI) agent, which delegates to specialist
agents. The agents draw on a knowledge base, a literature library, and a skill
library and are evaluated by BrainPilotBench. The Graph of Trace along the
bottom records every major step, with its nodes displaying per-agent status.}
\label{fig:system}
\end{figure}

BrainPilot is a system for
accelerating brain science research by coordinating domain knowledge,
analytical skills, and specialist agents within a traceable workflow implemented on PI Agent\footnote{\url{https://github.com/earendil-works/pi}}.
It combines three
elements to serve as a agent-verified research assistant: specialist agents that
carry out the research (Section~\ref{sec:agents}), curated domain knowledge that
grounds those agents in brain science (Section~\ref{sec:knowledge}), and a
Graph of Trace that records the entire process in an auditable form
(Section~\ref{sec:got}).
Figure~\ref{fig:system} shows how these parts are integrated into one system.

\subsection{Multi-agent system and agent roles}
\label{sec:agents}

BrainPilot carries out a research task with a team of specialist agents led by
the PI agent. The user states a goal to the PI, which decomposes it, delegates
subtasks to the specialist agents suited to each part, and integrates their
results into the final output. The user can view and inspect the work of every
agent. The default team has six agents (Table~\ref{tab:agents}). Five carry out the
research, while the Auditor makes the workflow trustworthy. Each agent is
defined with three parts: a shared system constitution establishes rules common
to all agents; a role definition specifies the agent's mission, working style,
and available tools; and a set of skills supplies its modular capabilities.
All agents share a messaging contract for tasks and replies, a self-recording
contract that logs their tangible outputs to the Graph of Trace.

\begin{table}[t]
\caption{The default agent team. Five agents carry out the research; the
Auditor checks the team's output before delivery.}
\label{tab:agents}
\begin{center}
\begin{tabular}{@{}>{\raggedright\arraybackslash}p{2.6cm}>{\raggedright\arraybackslash}p{10.2cm}@{}}
\multicolumn{1}{c}{\bf AGENT} & \multicolumn{1}{c}{\bf RESPONSIBILITY} \\ \hline \\
Principal (PI) & Coordinates the task, clarifies the user's intent, delegates to specialist agents, synthesizes their results, and routes deliverables through the Writer and Auditor before returning a final answer. The only agent the user instructs directly. \\[1.5pt]
Librarian & Surveys the literature and provides knowledge grounding, delivering a structured packet of Key Findings, Knowledge Gaps, Suggested Hypotheses, and References. \\[1.5pt]
Experimentalist & Designs the scientific procedure: operationalization, control design, sample planning, step-by-step protocol, and preregistered analysis plan. \\[1.5pt]
Engineer & Translates a design into reproducible code, runs it, and reports exactly what was executed and produced. \\[1.5pt]
Writer & Drafts and polishes scientific documents from the specialist agents' handoff packets, producing the report that the Auditor will check. \\[1.5pt]
Auditor & Independently audits each draft for fabrication before delivery, classifying numerical, file, and citation claims against the evidence present in the session. \\
\end{tabular}
\end{center}
\vspace{-3mm}
\end{table}

Each specialist agent combines a role-specific cognitive style with a small set
of hands-on tools. The Librarian emphasizes inductive synthesis and critical
evaluation, working exclusively through search and read tools. The
Experimentalist applies operational, control, and measurement thinking to
produce design documents, delegating substantial
implementation to the Engineer. The Engineer prioritizes precision and
reproducibility and uses file and shell tools to write, run, and report code in
the session workspace. The Writer focuses on clarity, precision, and audience
awareness when assembling final documents in that workspace. The PI delegates
work that requires domain expertise or more than a few minutes, while handling
lightweight framing, synthesis, and decisions directly. It can also perform
small inspections and produce short artifacts for simple requests.

Agents acquire task-specific abilities through modular capability units called
skills. Access to them is mediated by a skill-router tool. An agent queries the
router with keywords for its current task and reads the returned skills on
demand. The PI, Librarian,
Experimentalist, Engineer, and Writer all share this tool and follow the same
\emph{skills-first} preflight: before committing to a nontrivial methodology,
the agent queries for a matching skill and, when one exists, follows its
guidance as the methodological basis. This keeps each agent's working context
short and anchors its choices in practice.

The \textbf{Auditor} agent has a deliberately narrow role to check if the output of other agents contain errors which influence the scientifc results. Language-model agents can produce fluent claims unsupported by
anything in the session, and in a scientific setting such a claim constitutes
a substantive failure. To guard against this, the PI routes specialist-agent
output through the appropriate review stages before delivery. It first routes
deliverables that need report-like structure through the Writer, then sends the
resulting draft to the Auditor for independent verification. The Auditor asks questions claim in the draft: is it backed by evidence in the
session's own workspace? It classifies numerical claims, file or artifact
references, and external citations as confirmed, unverified, or disputed by
checking files produced by the other agents. When evidence is
ambiguous, it may ask the originating agent for a specific pointer. Each audit
is saved as a timestamped report in the session workspace, and a short message
notifies the PI of its location. This integrates verification into long-horizon workflow and relieves the user of that burden.

\begin{figure}[t]
\centering
\includegraphics[width=\linewidth]{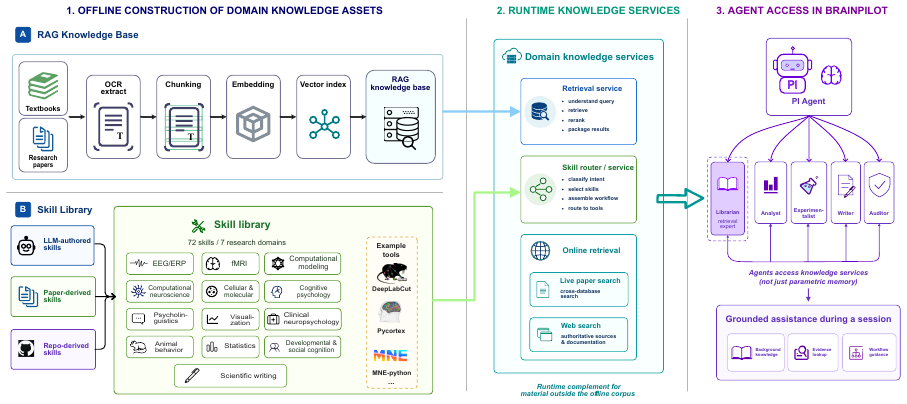}
\caption{\textbf{Knowledge-base construction.} Neuroscience textbooks and papers
are converted to text using OCR, split into overlapping passages, and embedded
in a unified vector index. At query time, an agent's question is embedded,
candidate passages are retrieved by vector similarity, and a cross-encoder
reranker reorders them before they are returned as evidence.}
\label{fig:kb}
\vspace{-3mm}
\end{figure}

\subsection{Domain knowledge}
\label{sec:knowledge}

BrainPilot grounds its agents in brain science through two knowledge assets
built offline: a retrieval knowledge base and a
skill library. Both are exposed to the agents as services, allowing them to
combine curated domain knowledge with their parametric knowledge.

\begin{figure*}[t]
\centering
\includegraphics[width=\textwidth]{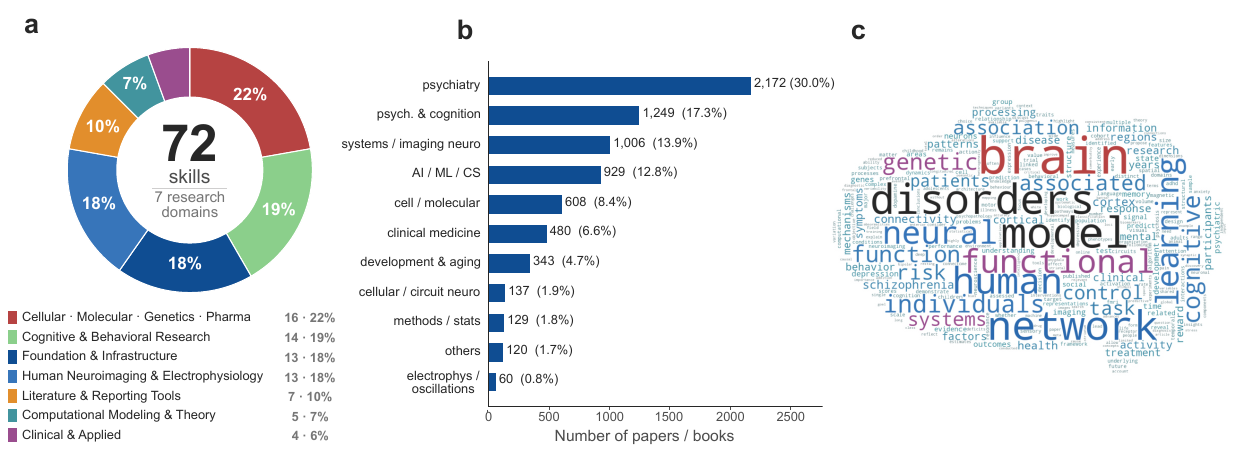}
\caption{\textbf{Composition of BrainPilot's domain-knowledge assets.} The
skill library comprises 72 skills across seven research domains. The knowledge
base contains 7{,}233 items assigned to 11 disciplinary groups. Its vocabulary
is summarized from 377{,}868 content tokens in paper titles and abstracts; the
400 most frequent retained terms are shown.}
\label{fig:corpus}
\vspace{-3mm}
\end{figure*}

\paragraph{Knowledge base.} The retrieval knowledge base is a single unified vector index built from
open-source neuroscience textbooks and the research literature, 7{,}233 indexed books and papers in total.\footnote{All resources indexed were obtained with the respective permission of rights holders.} The corpus spans 11 disciplinary groups. Psychiatry accounts for the largest share (2{,}172 items; 30.0\%), followed by psychology and cognition (1{,}249; 17.3\%), systems and imaging neuroscience (1{,}006; 13.9\%), and AI, machine learning, and computer science (929; 12.8\%). Paper titles and abstracts collectively contain 377{,}868 content tokens. Figure~\ref{fig:corpus} presents the 400 most frequent terms after excluding common scientific-prose filler.

The construction of the knowledge base proceeds in three stages (Figure~\ref{fig:kb}). Source documents are first converted to text with
DeepSeek-OCR~\citep{wei2025deepseek}.\footnote{\url{https://huggingface.co/deepseek-ai/DeepSeek-OCR}}
The text is then split into overlapping passages (1{,}500-character chunks with
a 200-character overlap), and each passage is embedded into a
1{,}024-dimensional vector with the dense retrieval encoder BAAI
bge-m3~\citep{chen2024bgem3}.\footnote{\url{https://huggingface.co/BAAI/bge-m3}}
At query time, an agent's question is embedded and matched against the index in
a two-stage retrieval procedure. Candidate passages are first selected by
vector similarity and then reordered by relevance using a cross-encoder
reranker from the same model family, BAAI
bge-reranker-v2-m3.\footnote{\url{https://huggingface.co/BAAI/bge-reranker-v2-m3}}
This gives the Librarian and the rest of the team access to textbook background
and published findings as evidence. The same pipeline can process newly added literature
and allow the corpus to grow for use with any agent applications.

\paragraph{Skill library and routing.} 
A skill is a self-contained,
reusable capability unit that may be descriptive or executable. Descriptive
skills record a method and when to apply it, while executable skills carry
code, such as a preprocessing routine derived from an established tool. The skill library contains specialized capabilities made available to agents. It currently holds 72
skills grouped into 7 broader research
domains: cellular, molecular, genetics, and pharmacology; cognitive and
behavioral research; foundation and infrastructure; human neuroimaging and
electrophysiology; computational modeling and theory; literature and reporting
tools; and clinical and applied research. 
We also design a skill router for a growingly number of skills exposed to agent's contexts. 
The router provides a two-stage interface between agents and the skill
library. An agent
first searches the library using task-relevant keywords and receives a compact
set of candidate skills and then loads the complete instructions for a selected
skill only when that capability is needed. The PI, Librarian, Experimentalist,
Engineer, and Writer share this interface and apply it as a skills-first
preflight before adopting a nontrivial method. Separating skill discovery from
skill loading limits context overhead while making searches and complete skill
loads explicit and measurable during execution.

Skills are drawn from three
complementary sources: an
LLM-driven authoring pipeline that produces general-purpose skills from a
specification, a paper-derived pipeline that distills methodological
prescriptions from brain science publications, and a tool-derived pipeline
that turns established analysis software into practical guides. The library
currently integrates skills derived from widely used neuroscience analysis
tools, including DeepLabCut for animal pose estimation
\citep{mathis2018deeplabcut}, MNE-Python for processing human neurophysiological data\footnote{\url{https://github.com/mne-tools/mne-python}}, fMRIPrep for preprocessing functional magnetic resonance imaging (fMRI) data\footnote{\url{https://github.com/nipreps/fmriprep}}, netneurotools for network neuroscience~\citep{Liu2025}, pycortex for visualizing volumetric neuroimaging data on cortical surfaces~\citep{10.3389/fninf.2015.00023}, and SpikeInterface for spike sorting~\citep{buccino2020spikeinterface}. 
The library
also includes meta-skills that extend the construction process for new sources:
\texttt{paper-to-skill} and \texttt{repo-to-skill} distill methods from
publications and software, while \texttt{workflow-skill-creator}\footnote{\url{https://github.com/google-deepmind/science-skills}} packages
completed procedures. These meta-skills provide a common, extensible workflow
for turning new methodological knowledge into reusable capability units while
retaining an explicit verification step.

\subsection{Graph of Trace}
\label{sec:got}

The third element, the Graph of Trace (GoT), makes the workflow inspectable. It
is implemented as a tool available to the agents
(Figure~\ref{fig:system}). After completing a subtask, each major agent calls the
GoT tool and reports what it produced. The tool builds the graph from these
reports, keeping provenance separate from how any individual agent runs. It
converts each report into a schema-validated node containing a
description, references to the artifacts produced, and the identifiers of its
parent steps. It then infers dependency edges that encode causal or logical
progression beyond temporal order. Nodes are appended to an
append-only record that forbids retrospective edits, cycles, and orphaned
references, so the graph grows monotonically into an explicit directed acyclic
graph of the workflow. A frontend renders this graph in real time and lets the
user inspect any node's metadata while execution continues, allowing the
research trajectory to be followed, reviewed, and corrected before an error
propagates. The design and evaluation of the Graph of Trace is demonstrated in ~\citep{gao2026got}.

\begin{figure}[t]
\centering
\includegraphics[width=\linewidth]{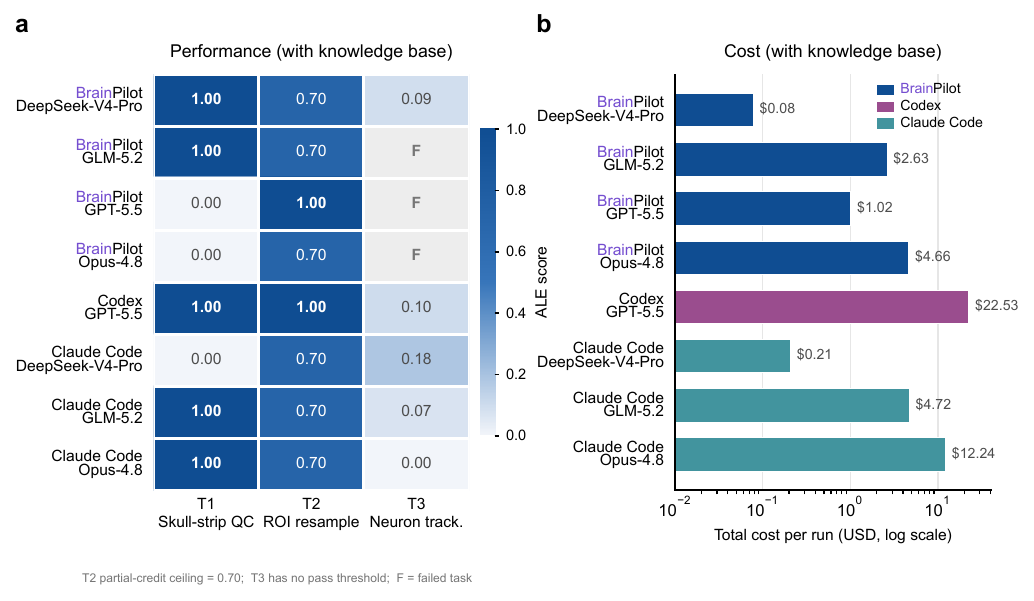}
\vspace{-3mm}
\caption{\textbf{\bpwordmark{} on Agents' Last Exam: performance and cost.}
Eight harness~$\times$~backbone configurations on the three brain science tasks
(T1 skull-stripping QC, T2 ROI resampling, T3 \emph{C.\ elegans} neuron
tracking), with domain knowledge enabled. \textbf{(a)} ALE grader scores;
``F'' marks a failed task with no gradable deliverable. \textbf{(b)} Total monetary
cost (log scale), summed over the three tasks and spanning more than two orders
of magnitude. Rows share the same order in both panels; ALE's own graders
scored every deliverable against withheld references in single runs.}
\label{fig:ale}
\end{figure}

\section{Evaluation}
\label{sec:evaluation}

For a full quantitative evaluation of BrainPilot, we conduct three complementary evaluations. First, we
run BrainPilot on Agents' Last Exam (ALE), a
recent benchmark for AI agents on long-horizon, real-world
professional workflows~\citep{ale2026}.\footnote{\url{https://agents-last-exam.org}}
(Section~\ref{sec:ale}), which scores its outputs against held-out references.
Second, we introduce \textbf{BrainPilotBench-v0} (Section~\ref{sec:benchmark}), a preliminary benchmark for agentic research in brain science and
report results spanning seven harness $\times$ model configurations, two domain-
knowledge conditions, and four tasks.

\subsection{Evaluation on Agents' Last Exam}
\label{sec:ale}

We run BrainPilot on Agents' Last Exam (ALE), a benchmark for AI agents on long-horizon, real-world
professional workflows~\citep{ale2026}.
ALE covers more than 1{,}000 tasks across 55 subfields grouped into 13 industry
clusters. We selected the three ALE tasks that fall in brain science: two
experimental-psychology and neuroimaging tasks (a skull-stripping
quality-control decision and a region-of-interest mask resampling) and onef
computational-neuroscience task (whole-brain neuron tracking in \emph{C.\
elegans}). Each task is conducted under three agent harnesses---BrainPilot, Codex,
and Claude Code--- driven open-source or state-of-the-art backbone LLMs(\texttt{deepseek-v4-pro}, \texttt{glm-5.2}, \texttt{claude-opus-4.8}, and \texttt{gpt-5.5}). We evaluated two conditions, with the domain knowledge in Section~\ref{sec:knowledge} either disabled or enabled, and report results of single runs.

\paragraph{Results.} Figure~\ref{fig:ale} summarizes performance and cost, with
exact per-task scores reported in Table~\ref{tab:ale}. Across matched backbones when domain knowledge enabled, BrainPilot completed the runs in 8--63\% of the time
and at 5--56\% of the cost required by Codex or Claude Code. T2 performance was
identical between harnesses for every matched backbone; with
\texttt{gpt-5.5}, both BrainPilot and Codex reached full credit ($1.00$) when
domain knowledge was enabled. T1 showed no consistent winner: BrainPilot was
more robust with \texttt{deepseek-v4-pro}, whereas Claude Code retained full
credit with \texttt{claude-opus-4.8}; the outcome with \texttt{gpt-5.5}
reversed after domain knowledge was introduced. The clearest limitation
appeared on the open-ended neuron-tracking task (T3). Thus, these single-run results show a substantial time and cost
advantage for BrainPilot, alongside lower reliability on the most open-ended
task.

\begin{table}[t]
\centering
\caption{Agents' Last Exam scores, runtime, and cost by
harness~$\times$~backbone. Within each task, \dkoff{} and \dkon{} denote domain
knowledge disabled and enabled, respectively (T1 skull-stripping QC, T2 ROI
resampling, T3 \emph{C.\ elegans} tracking; Appendix~\ref{app:ale-tasks}). Runtime
and cost correspond to \dkon{}. Green marks full credit ($1.00$)
on T1 and T2; T3 has no pass threshold. ``F'' marks a failed task with no
gradable deliverable.
Each score reports a single run graded against a withheld reference.}
\label{tab:ale}
\label{tab:ale-cost}
\scriptsize
\setlength{\tabcolsep}{3.2pt}
\renewcommand{\arraystretch}{1.08}
\begin{tabular}{@{}l cc cc cc cc@{}}
\toprule
\multicolumn{1}{c}{\multirow{2}{*}{Harness $\times$ Backbone}} &
\multicolumn{2}{c}{T1} & \multicolumn{2}{c}{T2} &
\multicolumn{2}{c}{T3} &
\multicolumn{1}{c}{\multirow{2}{*}{Time}} &
\multicolumn{1}{c}{\multirow{2}{*}{Cost (\$)}} \\
\cmidrule(lr){2-3}\cmidrule(lr){4-5}\cmidrule(lr){6-7}
 & \dkoff & \dkon & \dkoff & \dkon & \dkoff & \dkon & & \\
\midrule
\multicolumn{9}{@{}l}{\textit{\bpwordmark}} \\
\raisebox{-0.16\height}{\includegraphics[height=0.92em]{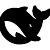}}\texttt{deepseek-v4-pro} & \textcolor{passgreen}{1.00} & \textcolor{passgreen}{1.00} & 0.70 & 0.70 & 0.08 & 0.09 & 17\,min & 0.08 \\
\raisebox{-0.16\height}{\includegraphics[height=0.92em]{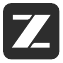}}\texttt{glm-5.2} & 0.00 & \textcolor{passgreen}{1.00} & 0.70 & 0.70 & 0.15 & F & 62\,min & 2.63 \\
\raisebox{-0.16\height}{\includegraphics[height=0.92em]{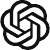}}\texttt{gpt-5.5} & \textcolor{passgreen}{1.00} & 0.00 & 0.70 & \textcolor{passgreen}{1.00} & F & F & 5\,min & 1.02 \\
\raisebox{-0.16\height}{\includegraphics[height=0.92em]{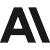}}\texttt{claude-opus-4.8} & \textcolor{passgreen}{1.00} & 0.00 & 0.70 & 0.70 & 0.07 & F & 18\,min & 4.66 \\
\midrule
\multicolumn{9}{@{}l}{\textit{Codex}} \\
\raisebox{-0.16\height}{\includegraphics[height=0.92em]{figs/model_icons/openai.pdf}}\texttt{gpt-5.5} & 0.00 & \textcolor{passgreen}{1.00} & 0.70 & \textcolor{passgreen}{1.00} & 0.10 & 0.10 & 63\,min & 22.53 \\
\midrule
\multicolumn{9}{@{}l}{\textit{Claude Code}} \\
\raisebox{-0.16\height}{\includegraphics[height=0.92em]{figs/model_icons/deepseek.pdf}}\texttt{deepseek-v4-pro} & \textcolor{passgreen}{1.00} & 0.00 & 0.70 & 0.70 & 0.03 & 0.18 & 44\,min & 0.21 \\
\raisebox{-0.16\height}{\includegraphics[height=0.92em]{figs/model_icons/glm.pdf}}\texttt{glm-5.2} & 0.00 & \textcolor{passgreen}{1.00} & 0.70 & 0.70 & 0.11 & 0.07 & 98\,min & 4.72 \\
\raisebox{-0.16\height}{\includegraphics[height=0.92em]{figs/model_icons/anthropic.pdf}}\texttt{claude-opus-4.8} & \textcolor{passgreen}{1.00} & \textcolor{passgreen}{1.00} & 0.70 & 0.70 & 0.09 & 0.00 & 48\,min & 12.24 \\
\bottomrule
\end{tabular}
\end{table}

\paragraph{Evaluation details.} Three details define the scope of these
results. First, ALE's tasks are written around desktop GUI software (3D Slicer,
FSLeyes, and a tracking application). BrainPilot used code in place of a
graphical interface, solved each task programmatically, and produced the same
required output files. This approach is permitted by the benchmark, whose
grading depends only on the delivered
artifacts.\footnote{Each task card scores on the deliverable: ``\emph{You only
pass if all required output files exist, are readable, and satisfy the
correctness checks against hidden reference data};'' the neuron-tracking card
states the work may be done ``\emph{either}'' through the GUI ``\emph{or}''
programmatically.} These results therefore probe the system's ability to
\emph{produce a correct computational deliverable}; operation through a
computer-use interface falls outside this evaluation. Second, each task's own
grader scored the deliverable against its withheld reference, while our adapter
orchestrated the run in place of ALE's hosted environment. The grader,
reference, and reveal-after-run ordering remain unchanged, making the numbers
grader-faithful re-scorings generated outside ALE's official hosted run. Third,
the backbone is a reasoning model with run-to-run variation, and we report one
unrepeated run per task and configuration.

\paragraph{Cost.}
Table~\ref{tab:ale-cost} reports the cost of each configuration with domain
knowledge enabled: agent wall-clock time, excluding offline grading, and the
corresponding monetary cost calculated from each backbone's official list
pricing. Both values are summed over the three tasks. Cost varies by more than
two orders of magnitude across backbones, from \$0.08 for BrainPilot with
\texttt{deepseek-v4-pro} to \$22.53 for Codex with \texttt{gpt-5.5}. The
open-ended neuron-tracking task (T3) accounts for most of this difference, as
its long trajectories dominate both time and token use. Read together with the
scores in Table~\ref{tab:ale}, the cheapest configuration is also fully
competitive on the deterministic tasks, pairing strong performance with the
lowest cost. Appendix~\ref{app:ale-tasks} provides the per-task breakdown
(Table~\ref{tab:ale-cost-detail}), and Table~\ref{tab:ale-cost-noKB} reports the
domain-knowledge-disabled cost ablation.

\subsection{BrainPilotBench-v0}
\label{sec:benchmark}

Whereas Agents' Last Exam is a general-purpose benchmark, we also
introduce \textbf{BrainPilotBench-v0}, a benchmark built
specifically for brain science agents. Like ALE, each task defines the input,
required deliverables, and a task-specific grader before execution; the grader
evaluates the submitted artifacts against a frozen reference or held-out labels
after the agent run~\citep{ale2026}. The initial suite contains four tasks
spanning calcium imaging, functional-connectivity fMRI, motor-imagery electroencephalogram(EEG), and
sleep EEG.

\paragraph{Task suite.} \emph{RSC place-cell analysis} provides multi-session
two-photon calcium activity and virtual-reality behavior from mice running on a
90-cm track~\citep{mao2020vision}. The agent performs place-cell screening,
population visualization, position decoding, trial-bin correlation, and
firing-rate analyses, and submits
a machine-readable summary, a scientific report, and figures. The grader checks
the place-cell ratio, decoding significance, and improvement over shuffled
decoding; visualization, trial-bin, and firing-rate outputs are retained as
diagnostics.

\emph{TOPS-fMRI} provides dynamic functional connectivity and pain ratings from an experimental tonic-pain cohort across 279 regions, yielding 38,781 edges ~\citep{lee2021painbiomarker}. The agent first trains a linear tonic-pain signature and submits its weight vector and intercept, manifest, NumPy-only inference program, edge-order validation, model card, and report. The grader then evaluates the submitted program on held-out clinical back-pain severity data, followed by two independent chronic-back-pain case–control cohorts from Japan and the UK. The final score combines condition-specific Pearson correlations for clinical pain severity with AUC values for patient–control discrimination in the two cohorts.

\emph{BCI Competition IV 2a} provides the labeled training session for each of
nine participants~\citep{tangermann2012bciiv}. The agent submits a PyTorch
\texttt{MIAgentModel} mapping $22\times512$ EEG samples to four motor-imagery
classes, together with a report.
Using fixed preprocessing and train--validation indices, the grader trains a
fresh model for each participant and evaluates 288 trials from that
participant's held-out evaluation session. The primary score is mean
Cohen's kappa across the nine participants; mean accuracy,
macro-F1, and participant-level accuracy and kappa are retained as diagnostics.

\emph{Sleep-EDF} provides single-channel Fpz--Cz recordings from subjects
0--15, using subjects 0--13 for training and 14--15 for
validation~\citep{kemp2000sleepedf,goldberger2000physionet}. The agent
submits a PyTorch \texttt{SleepAgentModel} mapping a 30-s, 100-Hz epoch to
Wake, N1, N2, N3, or REM, together with a report. The grader trains one shared
model under the fixed preprocessing and optimization protocol and evaluates it
on subjects 16--19. The primary score is Cohen's kappa on these held-out
subjects; accuracy, balanced accuracy, macro-F1, and
stage-specific recall provide diagnostic detail.

\paragraph{Artifact-based scoring.}
Each grader scores the submitted artifact produced during the agent run, not a
self-reported result. Task scores are normalized to $[0,1]$; a valid score of
zero is reported as $0.00$. ``F'' denotes a run for which no valid primary
score can be computed because required artifacts are missing or malformed, the
submission times out, or the scorer fails. It is not converted to numerical
zero. Appendix~\ref{app:brainpilotbench-scoring} gives the exact task-specific
definitions.

\paragraph{Isolation and integrity.} Task versions, prompts, artifact
contracts, preprocessing, data splits, scoring formulas, and content hashes are
fixed before execution. Agent runs receive only their prepared workspaces and
public task inputs. Scoring
stages evaluator inputs outside the agent workspace. Submitted code executes in
a no-network, non-root container with read-only code, model, and feature mounts;
only its prediction directory is writable. The trusted
task-owned grader reads held-out labels and computes the score only after the
submitted program exits.
Each task also passes a two-sided validation gate: a maintainer-authored Oracle
must produce the declared metric payload, whereas an empty submission must not.
This gate validates the scoring path rather than task performance. Before
official scoring, the submission bundle is checked against its artifact
contract; formal comparison bundles are then sealed before scoring. Missing
artifacts, malformed or non-finite predictions,
timeouts, and scorer failures produce explicit failed or unscored states rather
than numerical zeroes.

\begin{figure*}[t]
\centering
\includegraphics[width=0.98\textwidth]{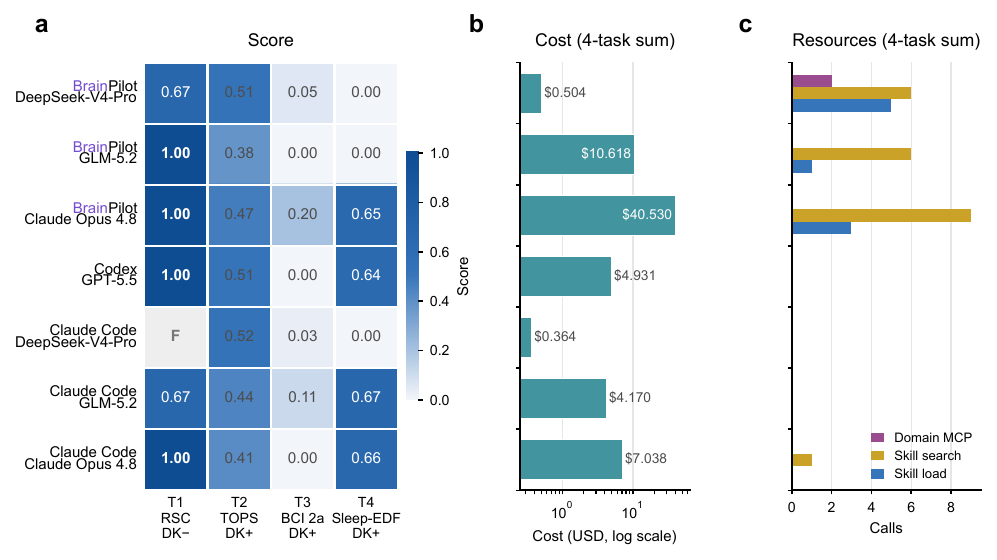}
\caption{\textbf{\bpbenchwordmark{} overview: performance, cost, and
domain-resource use.} \textbf{(a)} Scores across the four tasks:
T1 RSC place-cell analysis, T2 TOPS-fMRI, T3 BCI Competition IV 2a, and T4
Sleep-EDF. T1 reports the domain-knowledge-disabled score; T2--T4 report the
domain-knowledge-enabled score. ``F'' means that no valid primary score was
produced and is distinct from $0.00$. \textbf{(b)} Monetary cost summed over
all four domain-knowledge-enabled runs. \textbf{(c)} Domain-MCP calls, keyword
skill searches, and complete skill loads summed over all four tasks with domain
knowledge enabled. All results are reported by a single run.}
\label{fig:brainpilotbench-results}
\end{figure*}

\begin{table*}[t]
\centering
\caption{\bpbenchwordmark{} performance, runtime, and cost by
harness~$\times$~backbone. Within each task, \dkoff{} and \dkon{} denote domain
knowledge disabled and enabled, respectively. Time and cost are summed over all
four tasks with domain knowledge enabled. Costs are in US dollars using
official provider list prices verified on 16 July 2026, consistent with the
ALE analysis. For BrainPilot, RSC reports the higher-scoring of two runs, with
shorter runtime as the tie-breaker; all other cells report one formal run.
``F'' means that no valid primary score was produced and is not numerical zero.}
\label{tab:brainpilotbench-results}
\scriptsize
\setlength{\tabcolsep}{3.2pt}
\renewcommand{\arraystretch}{1.08}
\begin{tabular}{@{}l cc cc cc cc cc@{}}
\toprule
\multicolumn{1}{c}{\multirow{2}{*}{Harness $\times$ Backbone}} &
\multicolumn{2}{c}{RSC} &
\multicolumn{2}{c}{TOPS-fMRI} &
\multicolumn{2}{c}{BCI IV 2a} &
\multicolumn{2}{c}{Sleep-EDF} &
\multicolumn{1}{c}{\multirow{2}{*}{Time}} &
\multicolumn{1}{c}{\multirow{2}{*}{Cost (\$)}} \\
\cmidrule(lr){2-3}\cmidrule(lr){4-5}\cmidrule(lr){6-7}\cmidrule(lr){8-9}
 & \dkoff & \dkon & \dkoff & \dkon & \dkoff & \dkon & \dkoff & \dkon & & \\
\midrule
\multicolumn{11}{@{}l}{\textit{\bpwordmark}} \\
\raisebox{-0.16\height}{\includegraphics[height=0.92em]{figs/model_icons/deepseek.pdf}}\texttt{deepseek-v4-pro} & 0.67 & 0.33 & 0.49 & 0.51 & F & 0.05 & 0.66 & 0.00 & 80.2\,min & 0.504 \\
\raisebox{-0.16\height}{\includegraphics[height=0.92em]{figs/model_icons/glm.pdf}}\texttt{glm-5.2} & 1.00 & 0.33 & 0.45 & 0.38 & 0.00 & 0.00 & 0.00 & 0.00 & 117.8\,min & 10.618 \\
\raisebox{-0.16\height}{\includegraphics[height=0.92em]{figs/model_icons/anthropic.pdf}}\texttt{claude-opus-4.8} & 1.00 & 1.00 & 0.56 & 0.47 & 0.04 & 0.20 & 0.66 & 0.65 & 57.9\,min & 40.530 \\
\midrule
\multicolumn{11}{@{}l}{\textit{Codex}} \\
\raisebox{-0.16\height}{\includegraphics[height=0.92em]{figs/model_icons/openai.pdf}}\texttt{gpt-5.5} & 1.00 & 0.67 & 0.48 & 0.51 & 0.03 & 0.00 & 0.48 & 0.64 & 28.8\,min & 4.931 \\
\midrule
\multicolumn{11}{@{}l}{\textit{Claude Code}} \\
\raisebox{-0.16\height}{\includegraphics[height=0.92em]{figs/model_icons/deepseek.pdf}}\texttt{deepseek-v4-pro} & F & F & 0.35 & 0.52 & 0.00 & 0.03 & 0.65 & 0.00 & 91.9\,min & 0.364 \\
\raisebox{-0.16\height}{\includegraphics[height=0.92em]{figs/model_icons/glm.pdf}}\texttt{glm-5.2} & 0.67 & 1.00 & 0.49 & 0.44 & 0.00 & 0.11 & 0.70 & 0.67 & 57.6\,min & 4.170 \\
\raisebox{-0.16\height}{\includegraphics[height=0.92em]{figs/model_icons/anthropic.pdf}}\texttt{claude-opus-4.8} & 1.00 & 1.00 & 0.45 & 0.41 & 0.04 & 0.00 & 0.67 & 0.66 & 28.4\,min & 7.038 \\
\bottomrule
\end{tabular}
\end{table*}

\paragraph{BrainPilotBench results.} Figure~\ref{fig:brainpilotbench-results}
and Table~\ref{tab:brainpilotbench-results} summarize the 56 task cells, of
which 53 produced valid primary scores. With domain knowledge eabled, BrainPilot with
\texttt{claude-opus-4.8} tied the highest RSC score ($1.00$), achieved the
highest BCI IV 2a score ($0.20$), and outperformed the matched Claude Code
configuration on TOPS-fMRI ($0.47$ versus $0.41$), while scoring similarly on
Sleep-EDF ($0.65$ versus $0.66$). BrainPilot with
\texttt{deepseek-v4-pro} also produced a valid RSC result where Claude Code
failed, whereas BrainPilot with \texttt{glm-5.2} underperformed its matched
alternative across all four tasks. Figure~\ref{fig:brainpilotbench-results}
further shows that BrainPilot generally made broader use of the available
domain-knowledge tools. This greater tool use did not yield uniform score
improvements: across the 26 cells with valid scores in both conditions,
domain knowledge increased eight scores, reduced 14, and left four unchanged.

\paragraph{Runtime and cost.} Runtime and cost varied substantially across
backbones. BrainPilot with \texttt{deepseek-v4-pro} was faster than the matched
Claude Code configuration (80.2 versus 91.9 minutes), although slightly more
expensive (\$0.504 versus \$0.364). With \texttt{glm-5.2} and
\texttt{claude-opus-4.8}, BrainPilot required approximately twice the runtime
and cost 2.5 and 5.8 times as much, respectively, as the corresponding Claude
Code runs. The stronger results of BrainPilot with
\texttt{claude-opus-4.8} therefore came with a substantial cost premium.
Appendix~\ref{app:brainpilotbench} provides the per-task statistics.

\section{Case study}
\label{sec:casestudy}

We also carry out several 
end-to-end case studies in real scientific research scenarios, showing that BrainPilot can carry a multi-step analysis from scratch to academic and scientific results.

\subsection{Spatial coding in the retrosplenial cortex}
\label{sec:casestudy-spatial}

\begin{figure}[htbp]
\centering
\includegraphics[width=0.95\linewidth]{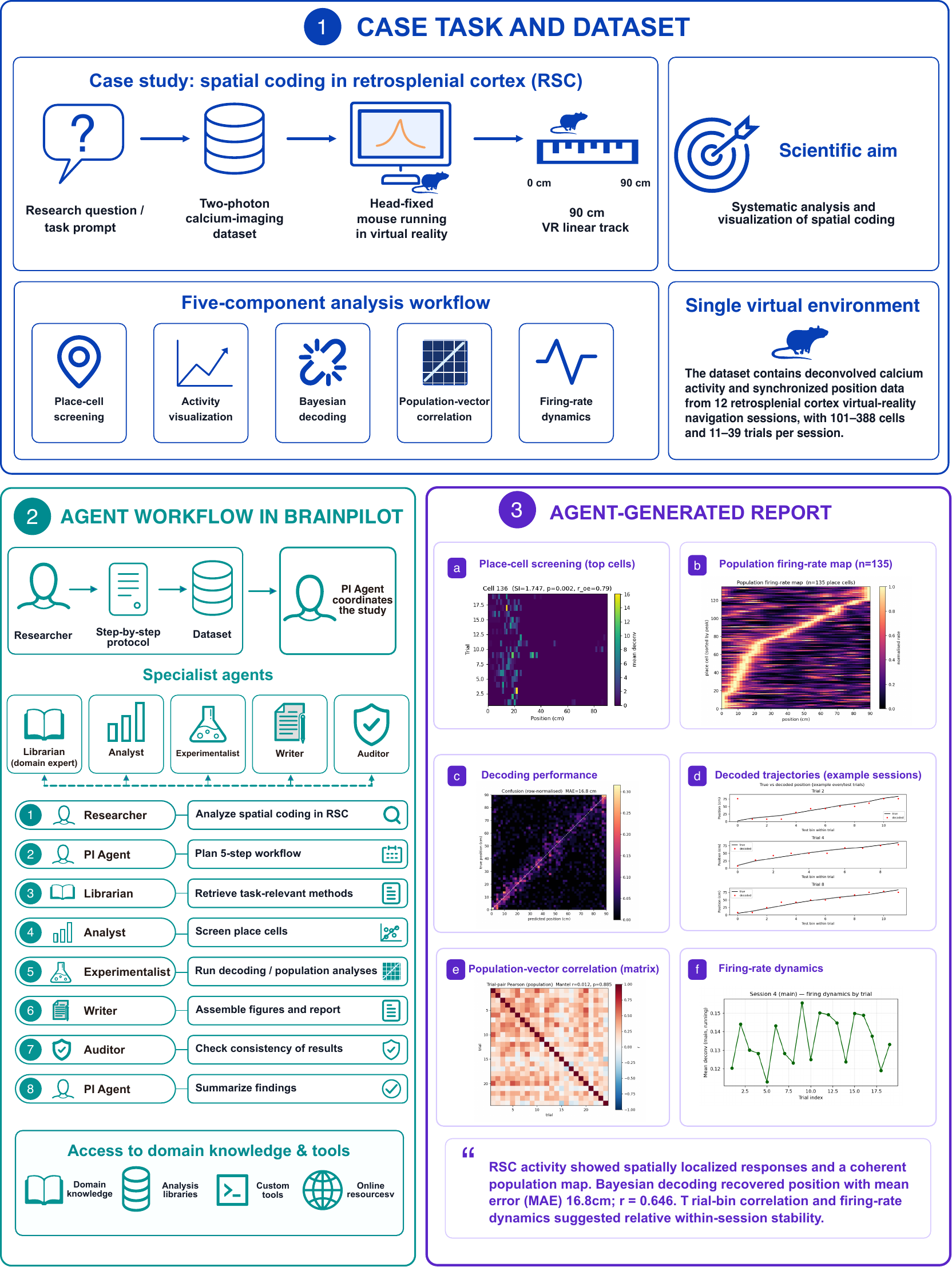}
\caption{\textbf{Case study: spatial coding in the retrosplenial cortex.} \textbf{(a)}
Representative single-cell activity heatmaps with spatial tuning curves.
\textbf{(b)} Population firing-rate map of detected place cells, sorted on odd
trials and rendered on held-out even trials. \textbf{(c)} Bayesian position
decoding: true-versus-decoded scatter and row-normalized confusion matrix.
\textbf{(d)} True and decoded positions across held-out test trials.
\textbf{(e)} Trial-bin population-vector correlation matrix. \textbf{(f)}
Peak-position drift and population mean peak rate across trials.}
\label{fig:casestudy}
\end{figure}

BrainPilot was evaluated on an end-to-end spatial-coding analysis in the retrosplenial cortex (RSC), a region implicated in spatial representation and navigation. This task is well suited to an agent because it requires coordinated decisions across dataset interpretation, neural–behavioral alignment, statistical screening, cross-validation, visualization, and scientific reporting, rather than the execution of a single predefined analysis function. The two-photon dataset
records RSC neural activity while head-fixed mice traverse a 90~cm virtual-reality
linear track~\citep{mao2020vision}. The researcher provided a high-level scientific objective and five requested analysis components:  \emph{place-cell screening}; \emph{single-cell and population visualization}; \emph{Bayesian position decoding}; \emph{trial-bin population-vector correlation}; and \emph{firing-rate dynamics}, but did not specify a complete implementation workflow. BrainPilot translated these requirements into an executable sequence of dataset inspection, session selection, temporal alignment, running-frame filtering, spatial-tuning analysis, held-out decoding, population-stability analysis, visualization, and report generation. The agent maintained a shared analysis state across these components. This reduced manual hand-offs and prevented inconsistent parameter choices across otherwise separate analyses. BrainPilot executed the step-by-step protocol (Appendix~\ref{app:case-prompts}) and produced the results
in Figure~\ref{fig:casestudy}.

Rather than returning isolated plots, BrainPilot organized the outputs into complementary levels of evidence. Representative heatmaps show localized activity bands recurring across trials
(Figure~\ref{fig:casestudy}a), consistent with place-field-like responses. The
cross-validated population map remains broadly diagonal on held-out even
trials when cells are ordered by preferences estimated from odd trials
(Figure~\ref{fig:casestudy}b). On held-out trials, the Bayesian decoder achieved
a mean absolute error of 16.8~cm and a true--decoded position correlation of
$r=0.646$ (Figure~\ref{fig:casestudy}c); reconstructed trajectories followed the
overall progression of true position with visible trial-level deviations
(Figure~\ref{fig:casestudy}d). The trial-bin correlation matrix was generally
positive beyond the unit diagonal, with stronger similarity among nearby bins
(Figure~\ref{fig:casestudy}e). Population mean peak rate remained within a
limited range across consecutive trials (Figure~\ref{fig:casestudy}f), further
supporting within-session stability.

Eventually, the five outputs provide complementary checks from single-cell tuning to population-level decoding and within-session dynamics. More importantly, the case evaluates whether an agent can convert a high-level neuroscience request into a coherent and reproducible analysis pipeline, resolve dataset-specific implementation details, preserve methodological consistency across dependent stages, and synthesize numerical results and figures into a scientific report.

\subsection{functional hierarchy in the mouse visual system}
\label{sec:casestudy-visual}

BrainPilot investigated functional hierarchy in the mouse visual system through
a literature-informed analysis of large-scale Neuropixels electrophysiology
data. Prior work has shown that spiking activity across mouse visual areas
reflects a functional hierarchy aligned with anatomical
hierarchy~\citep{siegle2021survey}. The Allen Brain Observatory Visual Coding
Neuropixels dataset contains approximately 135~GB of recordings from 58
sessions across multiple thalamic and cortical visual areas, including LGd,
VISp, VISl, VISrl, LP, VISal, VISpm, and VISam. The user provided predefined
anatomical hierarchy scores for these areas
(LGd $=-0.515$, VISp $=-0.357$, VISl $=-0.093$, VISrl $=-0.059$, LP $=0.105$,
VISal $=0.152$, VISpm $=0.327$, VISam $=0.441$).

\begin{figure}[t]
\centering
\includegraphics[width=0.95\linewidth]{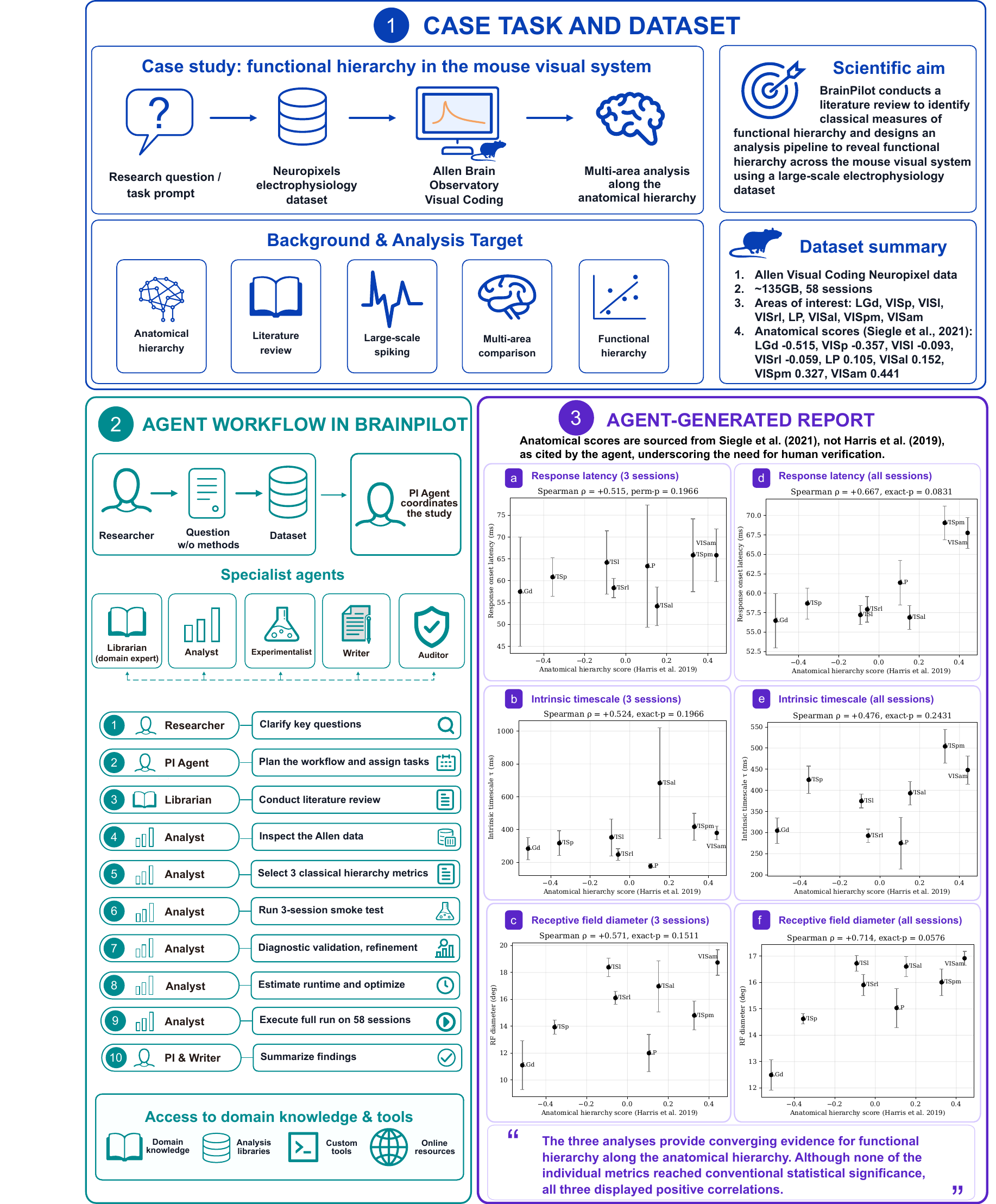}
\caption{\textbf{Case study: functional hierarchy in the mouse visual system.}
BrainPilot tested whether functional measures from spiking activity align with the
anatomical hierarchy of eight mouse visual areas. \textbf{(a--c)} Three-session
smoke-test outputs for response latency, intrinsic timescale, and receptive-field
diameter. \textbf{(d--f)} Full-run results for the same three metrics against the
anatomical hierarchy score.}
\label{fig:casestudy-visual}
\end{figure}

The prompt asked BrainPilot to identify classical physiological measures of
sensory hierarchy while forbidding reference to the motivating paper and its
code. The system had to justify each metric, predict its change along the
hierarchy, assess feasibility, and translate the literature into
quality-controlled spike processing, area-level estimates, and statistical
tests against the anatomical scores (Figure~\ref{fig:casestudy-visual}).

For each metric, session-level estimates were summarized within each visual area,
and significance was assessed by an exact permutation test over all $8!=40{,}320$
orderings of the eight area-level values against the anatomical hierarchy scores,
with the two-sided $p$-value defined as the fraction of permutations satisfying
$|\rho_{\mathrm{perm}}| \geq |\rho_{\mathrm{obs}}|$. Across the three metrics,
functional measures showed positive associations of differing strength with the
hierarchy. Flash-onset response latency increased with hierarchy score
(Figure~\ref{fig:casestudy-visual}d; Spearman $\rho=0.667$, exact permutation
$p=0.083$), consistent with progressively delayed processing along the visual
hierarchy. Intrinsic timescale showed a weaker positive trend
(Figure~\ref{fig:casestudy-visual}e; $\rho=0.476$, $p=0.243$). Receptive-field
diameter showed the largest correlation among the three
(Figure~\ref{fig:casestudy-visual}f; $\rho=0.714$, $p=0.058$), consistent with
increased spatial integration in higher visual areas. None of the three
correlations reaches conventional significance, but all are positive and align
with the hypothesized direction.

The prompt assigned the method selection, metric definitions, and statistical tests to the
agent. Its latency estimator used per-unit flash-aligned PSTHs and a
first-sustained-threshold definition, whereas prior work used median
time-to-first-spike; both recovered the same direction. An initial three-session
smoke test produced a negative latency--hierarchy correlation and implausibly
long values for LGd and VISp. The agent traced these values to noise
sensitivity, added a responsiveness criterion before latency estimation
(Figure~\ref{fig:casestudy-visual}a), and then scaled the corrected analysis to
the full dataset. This sequence retains the intermediate diagnosis and pipeline
revision alongside the final results.

\subsection{Functional-connectivity fMRI pain decoding}
\label{sec:case-fmri}

Sustained clinical pain is difficult to characterize objectively because it
unfolds over long timescales and is intertwined with cognitive and affective
processes. Experimental tonic pain provides a controlled model that may capture
neural features shared with ongoing clinical pain, making whole-brain functional
connectivity a promising basis for transferable biomarkers. A tonic-pain
functional-connectivity signature analysis therefore tests whether a signature
learned from experimentally induced tonic pain can generalize to clinical pain
severity and patient--control discrimination while retaining interpretable
network- and region-level structure~\citep{lee2021painbiomarker}.

\begin{figure}[htbp]
\centering
\includegraphics[width=0.95\linewidth]{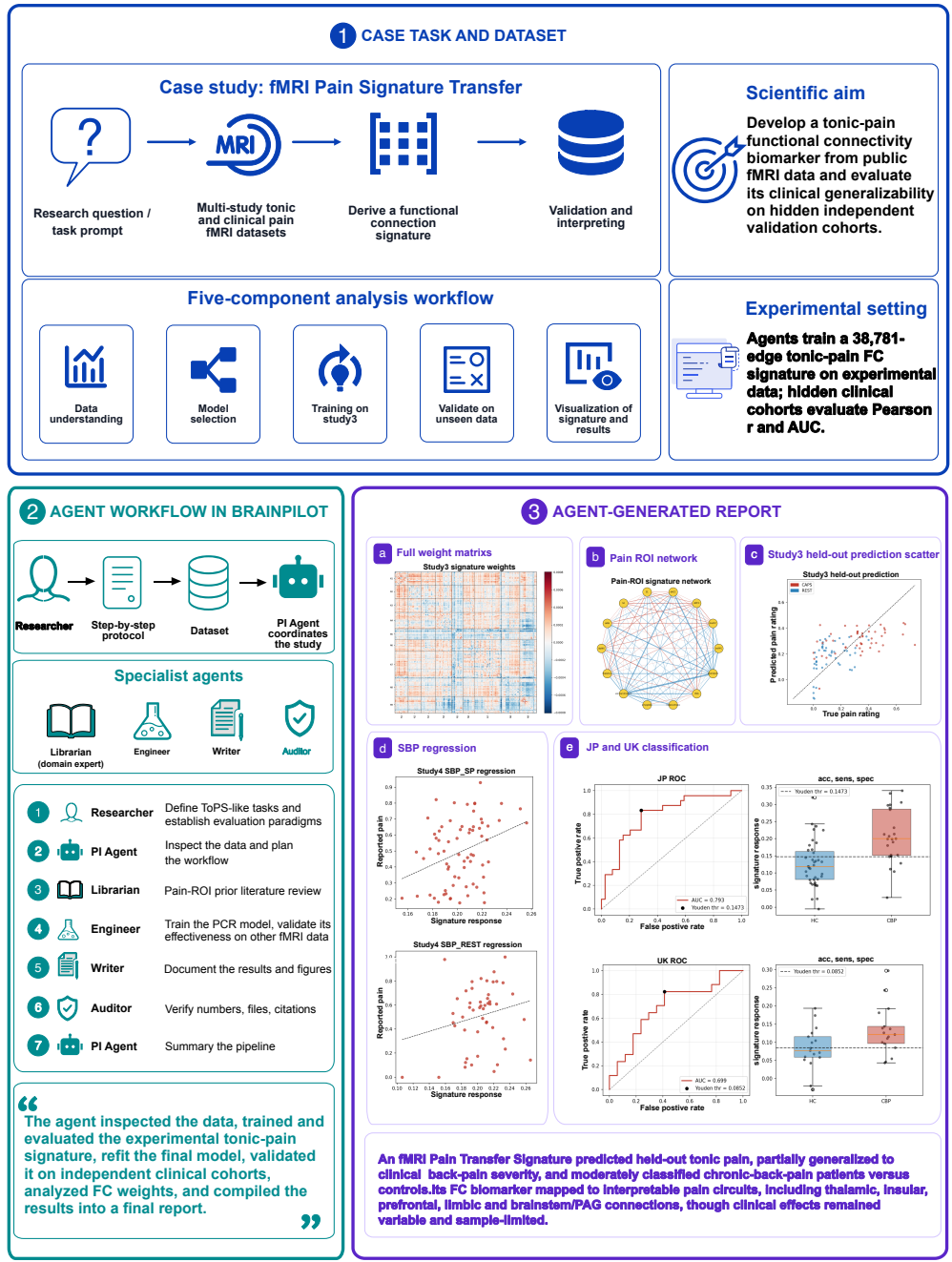}
\vspace{-4mm}
\caption{\textbf{Case study: an fMRI pain connectivity signature and its transfer.}
\textbf{(a)} Full FC weight matrix of the learned signature.
\textbf{(b)} Pain-region network of the signature weights.
\textbf{(c)} Held-out true-versus-predicted pain ratings in the experimental
tonic-pain cohort.
\textbf{(d)} External regression of signature response against pain severity in
the clinical back-pain cohort.
\textbf{(e)} Chronic-back-pain versus healthy-control classification in the
Japan and UK cohorts, shown using ROC curves and signature-response
distributions.}
\label{fig:case-fmri}
\end{figure}

To address this question, BrainPilot implemented an experimental-tonic-pain
connectivity-signature analysis that trained an FC signature on CAPS and REST
data and evaluated its transfer to independent clinical pain datasets. Each
sample comprised $38{,}781$ pairwise connections among 279 regions of interest.
The experimental tonic-pain cohort was first used for model training and
held-out internal evaluation. The frozen signature was then applied to pain
severity prediction across four subacute- and chronic-back-pain task conditions,
followed by chronic-back-pain classification in independent case--control
cohorts from Japan and the UK. The learned weights were finally mapped back to
FC space for network- and region-level interpretation
(Figure~\ref{fig:case-fmri}).

BrainPilot used a staged agent workflow. An engineering agent first inspected
the data, fixed the analysis configuration, trained and internally evaluated the
experimental tonic-pain signature, performed the external clinical validations,
and reconstructed the FC weights. A writing agent then reviewed the resulting
scripts, models, tables, and figures before drafting the report, ensuring that
each conclusion was tied to a saved output. Finally, an auditing agent
cross-checked the report against the underlying artifacts, identified minor
inconsistencies, and assessed the overall reliability of the completed analysis.

The experimental tonic-pain cohort contained 48 subjects per condition and 480
CAPS/REST subject--condition--time-bin samples. A fixed subject-level 80/20
split with seed 42 assigned 38 subjects to training and 10 to testing. The model
used standardization, 21-component PCA, and linear regression. After held-out
evaluation, it was refit on all experimental tonic-pain samples and converted
into a raw $38{,}781$-edge weight vector. The analysis excluded an erroneous
\texttt{pain\_avg\_dat} field from the experimental dataset and all-NaN FC
records from three clinical back-pain conditions.

On held-out subjects from the experimental tonic-pain cohort, the signature
predicted pain ratings with a pooled Pearson correlation of $r=0.5791$
($p=2.76\times10^{-10}$). The mean within-subject correlation was $r=0.6754$,
and CAPS responses exceeded REST responses in 9 of 10 subjects. The raw-space
weight vector reproduced the full PCR pipeline to numerical precision.

External validation provided partial evidence of clinical generalization. In
the subacute-back-pain spontaneous-pain condition, the signature significantly
predicted pain severity ($r=0.333$, $p=0.0048$), whereas the other three
clinical task and rest conditions showed positive but non-significant
associations. In the independent case--control datasets, the signature
distinguished chronic-back-pain patients from healthy controls with AUCs of
0.793 in Japan and 0.699 in the UK. Threshold-dependent classification metrics
were treated as descriptive because the thresholds were optimized within each
test cohort.

The signature was reconstructed as a symmetric $279\times279$ FC weight matrix,
and all 14 pain-related regions were mapped to 58 atlas parcels. Descriptive
weight patterns prominently involved the ventral striatum, brainstem, dACC,
aINS, and PAG. Interpretation was limited by the single split of the
experimental cohort, small clinical subgroups, weaker performance in the UK
cohort, imperfect edge-order verification, and the absence of inferential or
stability testing for individual FC weights.

\subsection{EEG motor-imagery decoding on BCI Competition IV 2a}
\label{sec:casestudy-eeg}

BrainPilot designed and evaluated a four-class EEG motor-imagery decoder on the
BNCI2014\_001 / BCI Competition IV 2a dataset~\citep{tangermann2012bciiv}, in
which subjects imagine movements of the left hand, right hand, both feet, or
tongue. From task constraints rather than an implementation template, the agent
had to produce an executable PyTorch model and choose how to encode frequency,
spatial, temporal, and small-sample priors.

The fixed preprocessing retained 22 EEG channels, applied a 4--38~Hz band-pass
filter, resampled to 128~Hz, and used a 0--4~s window, yielding inputs of
$(B,22,512)$ and four-class logits of $(B,4)$. Electrode activation maps and
temporal-kernel spectra tested motor-imagery priors: contralateral sensorimotor
activity near C3/C4 for hand imagery, midline activity near Cz for feet, and
more frontal-central activity for tongue imagery.

\begin{figure}[htbp]
\centering
\includegraphics[width=0.95\linewidth]{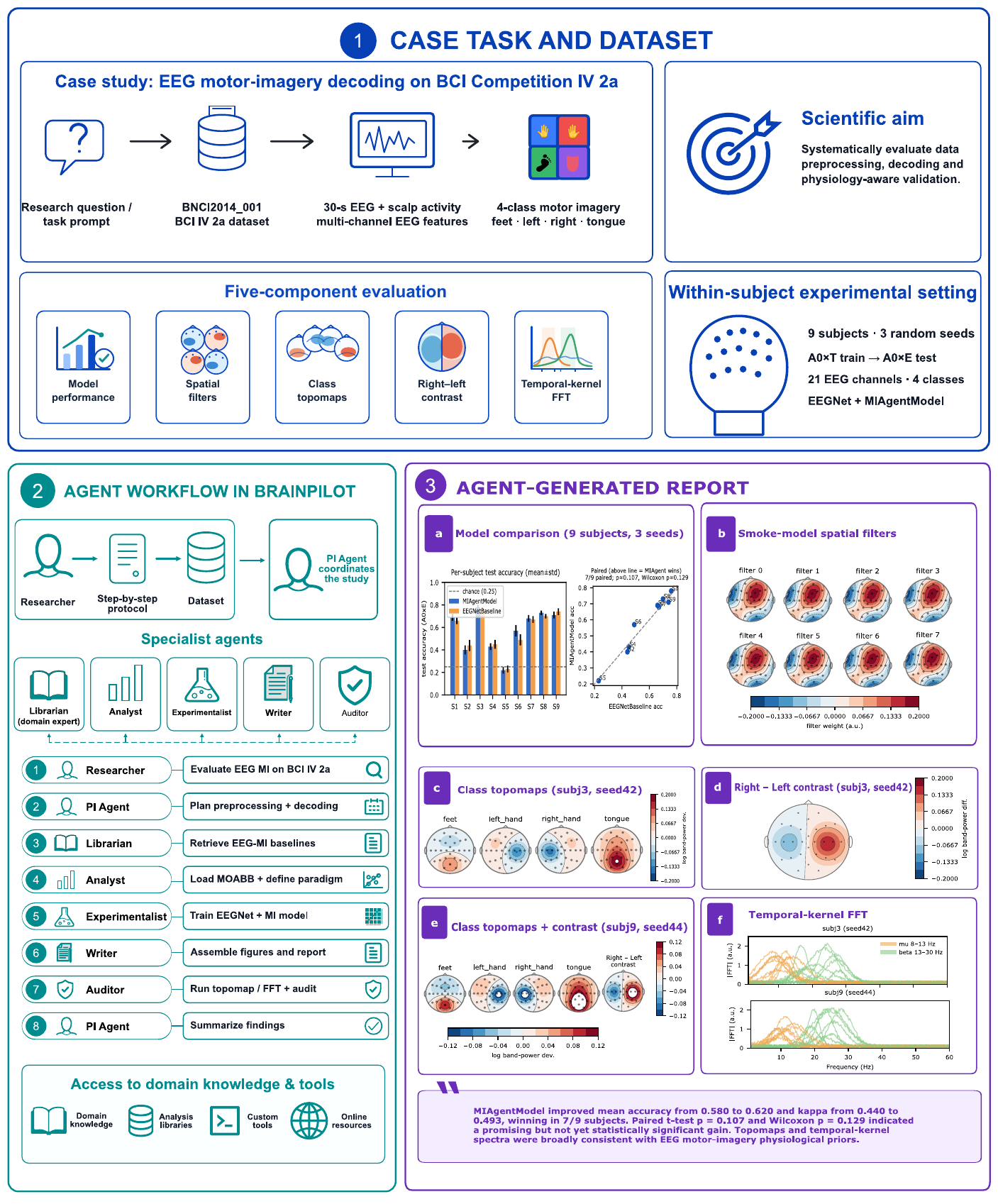}
\vspace{-3mm}
\caption{\textbf{Case study: EEG motor-imagery decoding on BCI Competition IV 2a.} \textbf{(a)} Model comparison (\texttt{MIAgentModel} vs.\ EEGNet baseline). \textbf{(b)} Learned spatial
filters as scalp topographies. \textbf{(c)} Class activation topomaps (subject 3).
\textbf{(d)} Right-minus-left contrast (subject 3). \textbf{(e)} Class topomaps and
contrast (subject 9). \textbf{(f)} Temporal-kernel FFT with Mu and Beta bands
highlighted.}
\label{fig:casestudy-eeg}
\end{figure}

The PI planned the workflow; the librarian reviewed EEGNet,
ShallowConvNet, and EEG-Conformer~\citep{lawhern2018eegnet,
schirrmeister2017deep, song2023eegconformer}; and the analyst prepared the
MOABB evaluation~\citep{jayaram2018moabb}. The experimentalist trained an
EEGNet baseline and \texttt{MIAgentModel}, then revised the latter to add a
temporal convolutional network (TCN) when the user requested an architecture
distinct from EEGNet. The writer assembled the report, and the auditor checked
the numerical claims (Figure~\ref{fig:casestudy-eeg}).

Within each subject, the A0xT session was used for training and A0xE for
testing, keeping the sessions separate. The final \texttt{MIAgentModel}
combined an EEGNet backbone with a TCN module, contained approximately 28.7k
parameters, mapped $(B,22,512)$ trials to $(B,4)$ logits, and passed import,
forward, backward, and training-interface checks.

Across 9 subjects and 3 random seeds, \texttt{MIAgentModel} outperformed the
EEGNet baseline in 7 of 9 subjects. Mean accuracy increased from 0.580 to
0.620, and Cohen's kappa increased from 0.440 to 0.493. The paired tests did
not reach conventional significance: the paired $t$-test gave $p=0.107$, and
the Wilcoxon signed-rank test gave $p=0.129$. The evidence therefore supports a
consistent positive trend rather than a statistically significant improvement
or a state-of-the-art claim.

Activation topomaps showed class-dependent sensorimotor patterns, with broadly
contralateral right-minus-left organization near C3/C4 and feet-related
activity closer to Cz. Temporal-kernel spectra covered the Mu and Beta bands.
These outputs support physiological plausibility but not a causal
interpretation of the learned filters.

The executable model showed a positive but nonsignificant trend over EEGNet.
Stronger claims require additional statistical evidence, label-shuffle checks,
and broader validation.

\begin{figure}[htbp]
\centering
\includegraphics[width=0.95\linewidth]{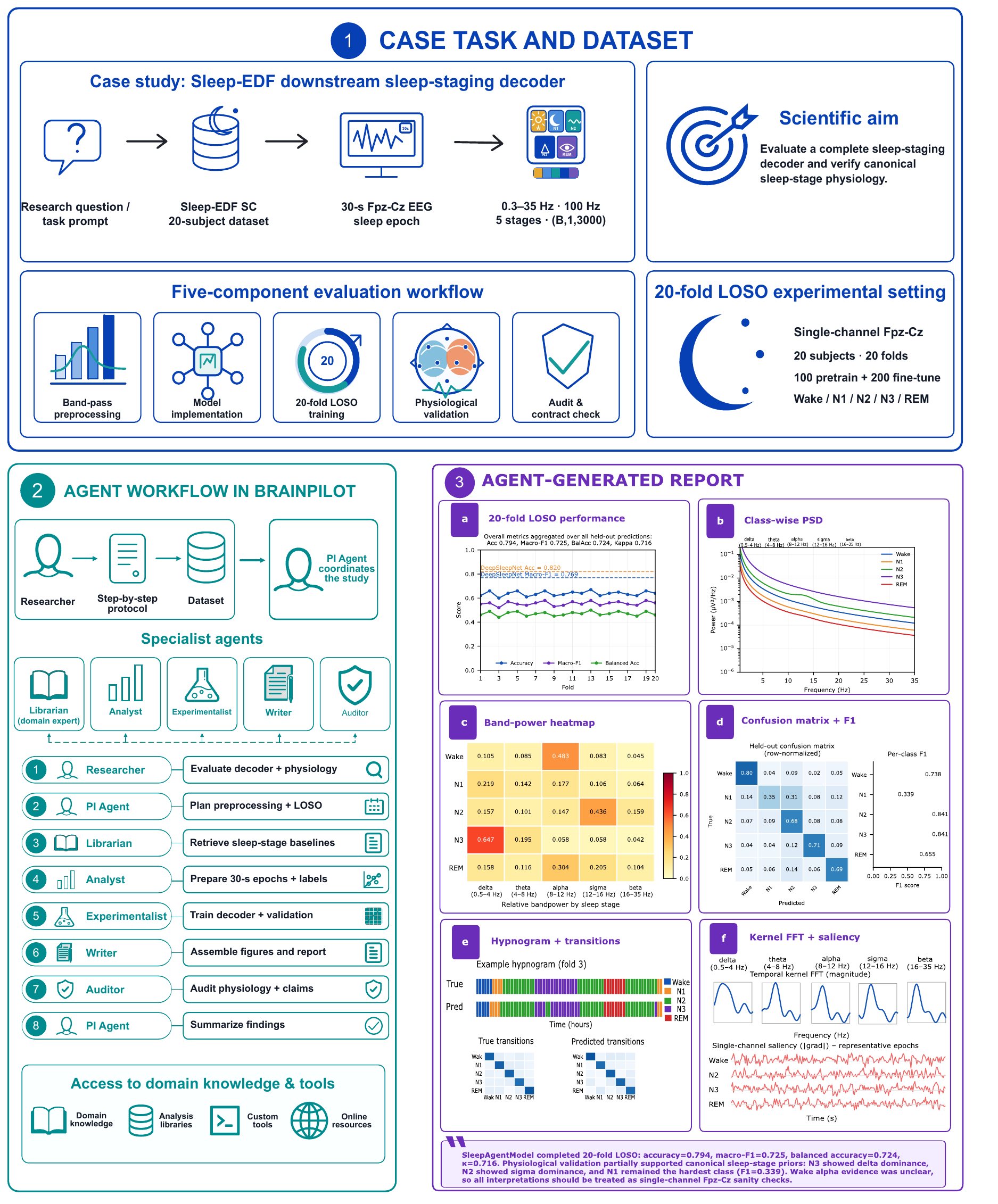}
\vspace{-3mm}
\caption{\textbf{Case study: Sleep-EDF downstream sleep-staging decoder with physiological validation.} \textbf{(a)} 20-fold leave-one-subject-out curves with DeepSleepNet reference lines.
\textbf{(b)} Class-wise power spectral density. \textbf{(c)} Relative band-power
heatmap across the five stages. \textbf{(d)} Held-out confusion matrix and
per-class F1. \textbf{(e)} True-versus-predicted hypnograms and transition
matrices. \textbf{(f)} Temporal-kernel FFT and saliency examples.}
\label{fig:casestudy-sleepedf-physio}
\end{figure}

\subsection{Sleep-EDF sleep staging with downstream decoding and physiological validation}
\label{sec:casestudy-sleepedf-physio}

BrainPilot performed five-class sleep staging on single-channel Fpz-Cz
recordings from 20 subjects in the Sleep-EDF Sleep Cassette
subset~\citep{kemp2000sleepedf, goldberger2000physionet}. Each 30-second epoch
was labeled Wake, N1, N2, N3, or REM. Unlike the multi-channel motor-imagery
case, this task emphasized long-window morphology, stage-specific frequency
structure, and hypnogram continuity through a runnable model, full
leave-one-subject-out evaluation.

The Sleep-EDF / DeepSleepNet-style protocol~\citep{supratak2017deepsleepnet}
sampled each epoch at 100~Hz, giving inputs of $(B,1,3000)$, and filtered signals
between 0.3 and 35~Hz. Movement and unknown stages were excluded, stages 3 and
4 were merged into N3, and Wake epochs were cropped around sleep. Each of 20
leave-one-subject-out folds trained on 19 subjects and tested on one, using 100
pretraining epochs followed by 200 fine-tuning epochs.

The resulting \texttt{SleepAgentModel} mapped $(B,1,3000)$ inputs to $(B,5)$
logits with 237,813 trainable parameters. It combined a temporal stem,
multi-scale branches spanning delta through beta morphology, a fusion block,
residual dilated temporal convolutions, and a five-class classifier.
Physiological validation included power spectra, band-power heatmaps,
temporal-kernel FFT, gradient saliency, hypnograms, and transition matrices
(Figure~\ref{fig:casestudy-sleepedf-physio}).

Across held-out predictions from all 20 subjects, \texttt{SleepAgentModel}
achieved 0.794 accuracy, 0.725 macro-F1, 0.724 balanced accuracy, and 0.716
Cohen's kappa. All four metrics exceeded majority-class and class-prior random
baselines but remained slightly below the DeepSleepNet
references~\citep{supratak2017deepsleepnet}.

Wake, N2, N3, and REM had F1 scores of 0.738, 0.841, 0.841, and 0.655,
respectively. N1 was the most difficult class ($F1=0.339$), with errors
concentrated in neighboring or transitional stages, consistent with its
ambiguity and with a nonuniform difficulty profile across stages.

Power spectra showed the highest delta power in N3 and sigma enhancement in N2,
consistent with slow-wave and spindle-related physiology. Temporal-kernel
spectra covered delta, theta/alpha, sigma, and higher-frequency components.
Gradient saliency identified locally relevant waveform segments but provided
only weak attribution evidence; hypnograms and transition matrices preserved
broad stage continuity while showing local errors.

Wake alpha power was not clearly dominant, and the single Fpz-Cz channel
precludes spatial or topographic interpretation. Stronger claims require a
reproduced DeepSleepNet baseline in the same codebase, additional datasets, and
more rigorous event-level analyses of spindles, K-complexes, and slow waves.

\section{Discussion}
\label{sec:discussion}

\paragraph{Benchmarking and evaluation.} Evaluating an agentic system for brain
science remains an open, demanding and resource-cost problem. The field offers few benchmarks
that measure a complete research workflow, with most focusing on isolated
knowledge or task accuracy. Building one realistic task requires
substantial expert time to source data, specify
the required artifacts and scoring contract, freeze the reference or held-out
data, and validate a task-specific grader. Among the more
than one thousand tasks in Agents' Last Exam, only three fall in neuroscience.
A field-wide evaluation therefore calls for a community effort. We treat
BrainPilotBench-v0 as a first framework whose scoring harness is already in
place. Adding a task requires its data, prompt, artifact contract, and grader,
and broader
coverage will extend the benchmark across more subdomains. \textbf{We release the methodology and first tasks openly and invite the
community to propose tasks, contribute data, define ground truth, and refine
scoring.}

\paragraph{Domain knowledge and laboratory experience.}
A field's working knowledge spans textbooks, research papers, analytical tools, hands-on experience, and laboratory-specific conventions. (1)~\textbf{Knowledge base:} BrainPilot organizes published knowledge into a retrieval corpus that can grow as new literature is processed through the same ingestion and indexing pipeline. (2)~\textbf{Skill library:} Skills record reusable methods together with their applicability boundaries and parameter choices. These distinctions are essential because the same experimental paradigm may require different analyses and parameters under different acquisition conditions, data properties, and research objectives. Meta-skills further convert papers, software, completed workflows, and expert knowledge into reviewable skill units, allowing the library to evolve through community contributions and providing a construction pattern applicable to other knowledge-intensive scientific domains. \textbf{We also call for the community to contribute more concrete and precise skills from lab experiences.}

\paragraph{Human-in-the-loop and scientific validity.}
BrainPilot supports expert participation through clarification questions, pre-execution discussion with the PI agent, and interfaces for inspecting intermediate actions and outputs. Beyond the end-to-end task completion emphasized in our case studies, researchers can refine the problem formulation, supply study-specific prior knowledge, and review methodological choices before execution, helping prevent unsuitable decisions from propagating through the workflow. \textbf{Meanwhile, researchers retain responsibility for validating BrainPilot's analyses, interpretations, and conclusions before using them to support scientific claims.} The Auditor incorporates an additional verification step by checking whether claims are supported by evidence in the session, while broader checks of methodological soundness, statistical validity, and scientific interpretation remain important directions for expert-centered verification.

\paragraph{Compositional workflows for research subtasks.} At present, a
demanding subtask such as an in-depth literature investigation or the drafting
of a academic paper is carried out by a single specialist agent and
reviewed by the PI. The community has developed mature workflows for exactly
these activities. Allowing a specialist agent to instantiate subworkflows and
subagents on demand would support more elaborate, task-appropriate, and
multi-stage procedures while preserving the same human-in-the-loop and
auditable structure. Composing established deep-research and academic-writing
pipelines into the existing PI--specialist--human architecture would deepen the
capabilities of each specialist agent.

\paragraph{Multimodal and cross-modal data integration.} Brain science advances
by integrating evidence across scales and modalities---electrophysiology,
two-photon and functional magnetic resonance imaging, EEG, behavior, and molecular data---and many
of the most important questions live precisely at the boundaries between them.
Our current case studies each operate within a single modality; extending
BrainPilot to jointly load, align, and reason over heterogeneous modalities, and
to draw cross-modal conclusions, would let the system address questions that
require evidence from multiple data types.

\section{Conclusion}
\label{sec:conclusion}

We presented BrainPilot, a human-in-the-loop multi-agent system that accelerates
brain science research while keeping the process ready for experts' verification. A principal
investigator (PI) agent coordinates specialist agents for literature grounding,
experimental design, execution, writing, and independent audit. Their work is
grounded in curated domain knowledge: a unified brain science knowledge base
containing 7{,}233 indexed items and a skill library of 72 reusable methodology
units across seven research domains. The Graph of
Trace provides an append-only, auditable record linking subgoals, tool use,
evidence, and claims, while a dedicated Auditor agent incorporates fabrication
checking into the method. This architecture allows researchers to follow,
inspect, and steer a study throughout its execution.

We evaluated BrainPilot on Agents' Last Exam, BrainPilotBench-v0 with four brain
science tasks, and end-to-end case
studies spanning systems neuroscience, mouse visual physiology,
functional-connectivity fMRI, and EEG decoding. Across these evaluations,
BrainPilot with an open-source backbone model achieved performance comparable
to state-of-the-art agent framework at a
lower monetary cost. We finally call again for the contribution of the community to a broad future of accelarating and automating brain discovery with agentic research.

\section*{{Acknowledgements}}

We sincerely thank the Prof. Anqi Wu from Georgia Institute of Technology, Prof. Sen Song from Tsinghua University, Dr. Yusi Chen from Institute of Neuroscience, Chinese Academy of Sciences for their detailed advice on BrainPilot. We thank all BrainPilot beta users for their thoughtful feedback on BrainPilot.

\bibliographystyle{iclr2025_conference}
\bibliography{references}

@article{gao2026got,
  title   = {Graph of Trace: Visualizing Execution Traces of Scientific Agents},
  author  = {Gao, Tianci and Li, Haoxuan and Li, Jianhe and Zhao, Tianxiang and Shi, Runze and Wang, Weiran and Wu, Zezhao and Mi, Lu},
  journal = {arXiv preprint arXiv:2606.15116},
  year    = {2026}
}

@article{boiko2023coscientist,
  title   = {Autonomous chemical research with large language models},
  author  = {Boiko, Daniil A. and MacKnight, Robert and Kline, Ben and Gomes, Gabe},
  journal = {Nature},
  volume  = {624},
  number  = {7992},
  pages   = {570--578},
  year    = {2023},
  doi     = {10.1038/s41586-023-06792-0}
}

@article{bran2024chemcrow,
  title   = {Augmenting large language models with chemistry tools},
  author  = {Bran, Andres M. and Cox, Sam and Schilter, Oliver and Baldassari, Carlo and White, Andrew D. and Schwaller, Philippe},
  journal = {Nature Machine Intelligence},
  volume  = {6},
  pages   = {525--535},
  year    = {2024},
  doi     = {10.1038/s42256-024-00832-8}
}

@article{huang2025biomni,
  title   = {Biomni: A General-Purpose Biomedical AI Agent},
  author  = {Huang, Kexin and Zhang, Serena and Wang, Hanchen and Qu, Yuanhao and Lu, Yingzhou and Roohani, Yusuf and Li, Ryan and Qiu, Lin and Li, Gavin and Zhang, Junze and Yin, Di and Marwaha, Shruti and Carter, Jennefer N. and Zhou, Xin and Wheeler, Matthew and Bernstein, Jonathan A. and Wang, Mengdi and He, Peng and Zhou, Jingtian and Snyder, Michael and Cong, Le and Regev, Aviv and Leskovec, Jure},
  journal = {bioRxiv},
  year    = {2025},
  doi     = {10.1101/2025.05.30.656746}
}

@article{jumper2021alphafold,
  title   = {Highly accurate protein structure prediction with {AlphaFold}},
  author  = {Jumper, John and Evans, Richard and Pritzel, Alexander and Green, Tim and Figurnov, Michael and Ronneberger, Olaf and Tunyasuvunakool, Kathryn and Bates, Russ and {\v{Z}}{\'\i}dek, Augustin and Potapenko, Anna and Bridgland, Alex and Meyer, Clemens and Kohl, Simon A. A. and Ballard, Andrew J. and Cowie, Andrew and Romera-Paredes, Bernardino and Nikolov, Stanislav and Jain, Rishub and Adler, Jonas and Back, Trevor and Petersen, Stig and Reiman, David and Clancy, Ellen and Zielinski, Michal and Steinegger, Martin and Pacholska, Michalina and Berghammer, Tamas and Bodenstein, Sebastian and Silver, David and Vinyals, Oriol and Senior, Andrew W. and Kavukcuoglu, Koray and Kohli, Pushmeet and Hassabis, Demis},
  journal = {Nature},
  volume  = {596},
  number  = {7873},
  pages   = {583--589},
  year    = {2021},
  doi     = {10.1038/s41586-021-03819-2}
}

@article{chen2026ethoclaw,
  title   = {{EthoClaw}: An Integrated AI Workflow Platform for Automated Analysis in Neuroethology},
  author  = {Chen, Ke and Chen, Ziming and Zheng, Dagang and Fang, Xiang and Liang, Jinghong and Li, Zhenyong and Chen, Yufeng and Zou, Jiemeng and Cai, Bingdong and Chen, Shanda and Huang, Kang},
  journal = {bioRxiv},
  year    = {2026},
  doi     = {10.64898/2026.03.25.714141}
}

@inproceedings{chen2024bgem3,
  title     = {{M3-Embedding}: Multi-Linguality, Multi-Functionality, Multi-Granularity Text Embeddings Through Self-Knowledge Distillation},
  author    = {Chen, Jianlv and Xiao, Shitao and Zhang, Peitian and Luo, Kun and Lian, Defu and Liu, Zheng},
  booktitle = {Findings of the Association for Computational Linguistics (ACL)},
  year      = {2024},
  note      = {arXiv:2402.03216}
}

@article{wei2025deepseek,
  title   = {{DeepSeek-OCR}: Contexts optical compression},
  author  = {Wei, Haoran and Sun, Yaofeng and Li, Yukun},
  journal = {arXiv preprint arXiv:2510.18234},
  year    = {2025}
}

@article{mathis2018deeplabcut,
  title   = {{DeepLabCut}: markerless pose estimation of user-defined body parts with deep learning},
  author  = {Mathis, Alexander and Mamidanna, Pranav and Cury, Kevin M. and Abe, Taiga and Murthy, Venkatesh N. and Mathis, Mackenzie W. and Bethge, Matthias},
  journal = {Nature Neuroscience},
  volume  = {21},
  number  = {9},
  pages   = {1281--1289},
  year    = {2018},
  doi     = {10.1038/s41593-018-0209-y}
}

@article{wang2026neuroclaw,
  title   = {{NeuroClaw} Technical Report},
  author  = {Wang, Cheng and He, Zhibin and Peng, Zhihao and Liu, Shengyuan and Hu, Yufan and Yang, Carl and He, Lifang and Sun, Lichao and Li, Xiang and Yuan, Yixuan},
  journal = {arXiv preprint arXiv:2604.24696},
  year    = {2026}
}

@inproceedings{hendrycks2021mmlu,
  title     = {Measuring Massive Multitask Language Understanding},
  author    = {Hendrycks, Dan and Burns, Collin and Basart, Steven and Zou, Andy and Mazeika, Mantas and Song, Dawn and Steinhardt, Jacob},
  booktitle = {International Conference on Learning Representations (ICLR)},
  year      = {2021},
  note      = {arXiv:2009.03300}
}

@inproceedings{chen2024scienceagentbench,
  title     = {{ScienceAgentBench}: Toward Rigorous Assessment of Language Agents for Data-Driven Scientific Discovery},
  author    = {Chen, Ziru and Chen, Shijie and Ning, Yuting and Zhang, Qianheng and Wang, Boshi and Yu, Botao and Li, Yifei and Liao, Zeyi and Wei, Chen and Lu, Zitong and Dey, Vishal and Xue, Mingyi and Baker, Frazier N. and Burns, Benjamin and Adu-Ampratwum, Daniel and Huang, Xuhui and Ning, Xia and Gao, Song and Su, Yu and Sun, Huan},
  booktitle = {International Conference on Learning Representations (ICLR)},
  year      = {2025},
  note      = {arXiv:2410.05080}
}

@inproceedings{majumder2024discoverybench,
  title     = {{DiscoveryBench}: Towards Data-Driven Discovery with Large Language Models},
  author    = {Majumder, Bodhisattwa Prasad and Surana, Harshit and Agarwal, Dhruv and Dalvi Mishra, Bhavana and Meena, Abhijeetsingh and Prakhar, Aryan and Vora, Tirth and Khot, Tushar and Sabharwal, Ashish and Clark, Peter},
  booktitle = {International Conference on Learning Representations (ICLR)},
  year      = {2025},
  note      = {arXiv:2407.01725}
}

@inproceedings{li2025autosdt,
  title     = {{AutoSDT}: Scaling Data-Driven Discovery Tasks Toward Open Co-Scientists},
  author    = {Li, Yifei and Moussa, Hanane Nour and Chen, Ziru and Chen, Shijie and Yu, Botao and Xue, Mingyi and Burns, Benjamin and Chiu, Tzu-Yao and Dey, Vishal and Lu, Zitong and Wei, Chen and Zhang, Qianheng and Zhang, Tianyu and Gao, Song and Huang, Xuhui and Ning, Xia and Ahmed, Nesreen K. and Payani, Ali and Sun, Huan},
  booktitle = {Proceedings of the Conference on Empirical Methods in Natural Language Processing (EMNLP)},
  year      = {2025},
  note      = {arXiv:2506.08140}
}

@article{luo2025brainbench,
  title   = {Large language models surpass human experts in predicting neuroscience results},
  author  = {Luo, Xiaoliang and Rechardt, Akilles and Sun, Guangzhi and Nejad, Kevin K. and Y{\'a}{\~n}ez, Felipe and others and Love, Bradley C.},
  journal = {Nature Human Behaviour},
  volume  = {9},
  number  = {2},
  pages   = {305--315},
  year    = {2025},
  doi     = {10.1038/s41562-024-02046-9}
}

@article{ale2026,
  title   = {Agents' Last Exam},
  author  = {Sun, Yiyou and Han, Xinyang and Zhang, Weichen and Pang, Yuanbo and Wang, Tianyu and Cao, Yuhan and Huang, Yixiao and Duroiu, Chris and Zhang, Haoyun and Lin, Jeffrey and others},
  journal = {arXiv preprint arXiv:2606.05405},
  year    = {2026}
}

@article{lu2024aiscientist,
  title   = {The {AI} Scientist: Towards Fully Automated Open-Ended Scientific Discovery},
  author  = {Lu, Chris and Lu, Cong and Lange, Robert Tjarko and Foerster, Jakob and Clune, Jeff and Ha, David},
  journal = {arXiv preprint arXiv:2408.06292},
  year    = {2024}
}

@article{lu2026endtoend,
  title   = {Towards end-to-end automation of {AI} research},
  author  = {Lu, Chris and Lu, Cong and Lange, Robert Tjarko and Yamada, Yutaro and Hu, Shengran and Foerster, Jakob and Ha, David and Clune, Jeff},
  journal = {Nature},
  volume  = {651},
  pages   = {914--919},
  year    = {2026}
}

@article{szymanski2023alab,
  title   = {An autonomous laboratory for the accelerated synthesis of inorganic materials},
  author  = {Szymanski, Nathan J. and Rendy, Bernardus and Fei, Yuxing and Kumar, Rishi E. and He, Tanjin and Milsted, David and McDermott, Matthew J. and Gallant, Max and Cubuk, Ekin Dogus and Merchant, Amil and Kim, Haegyeom and Jain, Anubhav and Bartel, Christopher J. and Persson, Kristin and Zeng, Yan and Ceder, Gerbrand},
  journal = {Nature},
  volume  = {624},
  pages   = {86--91},
  year    = {2023}
}

@article{ghareeb2026robin,
  title   = {A multi-agent system for automating scientific discovery},
  author  = {Ghareeb, Ali Essam and Chang, Benjamin and Mitchener, Ludovico and Yiu, Angela and Szostkiewicz, Caralyn J. and Shved, Dmytro and Gyimesi, Gavin J. and Laurent, Jon M. and Wright, Samantha M. and Razzak, Muhammed T. and White, Andrew D. and Finnemann, Silvia C. and Hinks, Michaela M. and Rodriques, Samuel G.},
  journal = {Nature},
  year    = {2026}
}

@article{gottweis2026coscientist,
  title   = {Accelerating scientific discovery with {Co-Scientist}},
  author  = {Gottweis, Juraj and Weng, Wei-Hung and Daryin, Alexander and Tu, Tao and Sirkovic, Petar and Myaskovsky, Artiom and Glowaty, Grzegorz and Weissenberger, Felix and Orlandi, Alessio and Popovici, Dan and Palepu, Anil and Rong, Keran and Tanno, Ryutaro and Saab, Khaled and Zhang, Fan and Blum, Jacob and Carroll, Andrew and Kulkarni, Kavita and Tomašev, Nenad and Zverinski, Dina and Rendulic, Ivor and Vedadi, Elahe and Hasler, Florian and Rimanic, Luka and Boia, Marina and Budiselic, Ivan and Feinstein, Ben and Bellaiche, Mathias and Sheffer, Tom and Freyberg, Jan and Ratcliff, Jeremy and Bertolli, Ottavia and Chou, Katherine and Hassidim, Avinatan and Gokturk, Burak and Vahdat, Amin and Guan, Yuan and Dhillon, Vikram and Vaishnav, Eeshit Dhaval and Lee, Byron and Costa, Tiago R. D. and Penadés, José R. and Peltz, Gary and Matias, Yossi and Manyika, James and Hassabis, Demis and Xu, Yunhan and Kohli, Pushmeet and Pawlosky, Annalisa and Karthikesalingam, Alan and Natarajan, Vivek},
  journal = {Nature},
  year    = {2026}
}

@article{zhou2026brainagent,
  title   = {{BrainAgent}: A Large Language Model-Driven Multi-Agent Framework for Autonomous Brain Signal Understanding},
  author  = {Zhou, Yangxuan and Zhao, Sha and Wang, Jiquan and Li, Shijian and Pan, Gang},
  journal = {arXiv preprint arXiv:2606.25400},
  year    = {2026}
}

@article{prystawski2026autopsych,
  title   = {auto-psych: Automating the science of mind using agent-driven theory discovery and experimentation},
  author  = {Prystawski, Ben and Mukherjee, Kushin and Wurgaft, Daniel and Nasvytis, Linas and Li, Michael Y. and Goodman, Noah D. and Frank, Michael C.},
  journal = {arXiv preprint arXiv:2606.26460},
  year    = {2026}
}

@article{xi2023rise,
  title   = {The Rise and Potential of Large Language Model Based Agents: A Survey},
  author  = {Xi, Zhiheng and Chen, Wenxiang and Guo, Xin and He, Wei and Ding, Yiwen and Hong, Boyang and Zhang, Ming and Wang, Junzhe and Jin, Senjie and Zhou, Enyu and Zheng, Rui and Fan, Xiaoran and Wang, Xiao and Xiong, Limao and Zhou, Yuhao and Wang, Weiran and Jiang, Changhao and Zou, Yicheng and Liu, Xiangyang and Yin, Zhangyue and Dou, Shihan and Weng, Rongxiang and Cheng, Wensen and Zhang, Qi and Qin, Wenjuan and Zheng, Yongyan and Qiu, Xipeng and Huang, Xuanjing and Gui, Tao},
  journal = {Science China Information Sciences},
  year    = {2024}
}

@inproceedings{guo2024multiagent,
  title     = {Large Language Model based Multi-Agents: A Survey of Progress and Challenges},
  author    = {Guo, Taicheng and Chen, Xiuying and Wang, Yaqi and Chang, Ruidi and Pei, Shichao and Chawla, Nitesh V. and Wiest, Olaf and Zhang, Xiangliang},
  booktitle = {Proceedings of the 33rd International Joint Conference on Artificial Intelligence (IJCAI)},
  year      = {2024},
  note      = {Also arXiv:2402.01680}
}

@article{wei2025agenticscience,
  title   = {From {AI} for Science to Agentic Science: A Survey on Autonomous Scientific Discovery},
  author  = {Wei, Jiaqi and Yang, Yuejin and Zhang, Xiang and Chen, Yuhan and Zhuang, Xiang and Gao, Zhangyang and Zhou, Dongzhan and Wang, Guangshuai and Gao, Zhiqiang and Cao, Juntai and Qiu, Zijie and Hu, Ming and Ma, Chenglong and Tang, Shixiang and He, Junjun and Song, Chunfeng and He, Xuming and Zhang, Qiang and You, Chenyu and Zheng, Shuangjia and Ding, Ning and Ouyang, Wanli and Dong, Nanqing and Cheng, Yu and Sun, Siqi and Bai, Lei and Zhou, Bowen},
  journal = {arXiv preprint arXiv:2508.14111},
  year    = {2025}
}

@article{zhang2025llmscimethod,
  title   = {Exploring the role of large language models in the scientific method: from hypothesis to discovery},
  author  = {Zhang, Yanbo and Khan, Sumeer A. and Mahmud, Adnan and Yang, Huck and Lavin, Alexander and Levin, Michael and Frey, Jeremy and Dunnmon, Jared and Evans, James and Bundy, Alan and Dzeroski, Saso and Tegner, Jesper and Zenil, Hector},
  journal = {npj Artificial Intelligence},
  volume  = {1},
  year    = {2025}
}

@article{liu2023trustworthy,
  title   = {Trustworthy {LLMs}: a Survey and Guideline for Evaluating Large Language Models' Alignment},
  author  = {Liu, Yang and Yao, Yuanshun and Ton, Jean-Francois and Zhang, Xiaoying and Guo, Ruocheng and Cheng, Hao and Klochkov, Yegor and Taufiq, Muhammad Faaiz and Li, Hang},
  journal = {arXiv preprint arXiv:2308.05374},
  year    = {2023}
}

@article{yu2025trustagents,
  title   = {A Survey on Trustworthy {LLM} Agents: Threats and Countermeasures},
  author  = {Yu, Miao and Meng, Fanci and Zhou, Xinyun and Wang, Shilong and Mao, Junyuan and Pang, Linsey and Chen, Tianlong and Wang, Kun and Li, Xinfeng and Zhang, Yongfeng and An, Bo and Wen, Qingsong},
  journal = {arXiv preprint arXiv:2503.09648},
  year    = {2025}
}

@article{lin2025agenthalluc,
  title   = {{LLM}-based Agents Suffer from Hallucinations: A Survey of Taxonomy, Methods, and Directions},
  author  = {Lin, Xixun and Ning, Yucheng and Zhang, Jingwen and Dong, Yan and Liu, Yilong and Wu, Yongxuan and Qi, Xiaohua and Sun, Nan and Shang, Yanmin and Wang, Kun and Cao, Pengfei and Wang, Qingyue and Zou, Lixin and Chen, Xu and Zhou, Chuan and Wu, Jia and Zhang, Peng and Wen, Qingsong and Pan, Shirui and Wang, Bin and Cao, Yanan and Chen, Kai and Hu, Songlin and Guo, Li},
  journal = {arXiv preprint arXiv:2509.18970},
  year    = {2025}
}

@article{vaccaro2024humanai,
  title   = {When combinations of humans and {AI} are useful: A systematic review and meta-analysis},
  author  = {Vaccaro, Michelle and Almaatouq, Abdullah and Malone, Thomas},
  journal = {Nature Human Behaviour},
  volume  = {8},
  year    = {2024}
}

@article{shao2025sciscigpt,
  title   = {{SciSciGPT}: advancing human--{AI} collaboration in the science of science},
  author  = {Shao, Erzhuo and Wang, Yifang and Qian, Yifan and Pan, Zhenyu and Liu, Han and Wang, Dashun},
  journal = {Nature Computational Science},
  volume  = {6},
  year    = {2025}
}

@article{siegle2021survey,
  title   = {Survey of spiking in the mouse visual system reveals functional hierarchy},
  author  = {Siegle, Joshua H. and Jia, Xiaoxuan and Durand, S{\'e}verine and Gale, Sam and Bennett, Corbett and Graddis, Nile and Heller, Greggory and Ramirez, Tamina K. and Choi, Hannah and Luviano, Jennifer A. and Groblewski, Peter A. and Ahmed, Ramakrishnan and Arkhipov, Anton and Bernard, Amy and Billeh, Yazan N. and Brown, Dillan and Buice, Michael A. and Cain, Nicholas and Caldejon, Shiella and Casal, Linzy and others},
  journal = {Nature},
  volume  = {592},
  number  = {7852},
  pages   = {86--92},
  year    = {2021}
}

@article{lee2021painbiomarker,
  title   = {A neuroimaging biomarker for sustained experimental and clinical pain},
  author  = {Lee, Jae-Joong and Kim, Hong Ji and {\v{C}}eko, Marta and Park, Bo-yong and Lee, Soo Ahn and Park, Hyunjin and Roy, Mathieu and Kim, Seong-Gi and Wager, Tor D. and Woo, Choong-Wan},
  journal = {Nature Medicine},
  volume  = {27},
  number  = {1},
  pages   = {174--182},
  year    = {2021},
  doi     = {10.1038/s41591-020-1142-7}
}

@article{tangermann2012bciiv,
  title   = {Review of the {BCI} Competition {IV}},
  author  = {Tangermann, Michael and M{\"u}ller, Klaus-Robert and Aertsen, Ad and Birbaumer, Niels and Braun, Christoph and Brunner, Clemens and Leeb, Robert and Mehring, Carsten and Miller, Kai J. and M{\"u}ller-Putz, Gernot R. and Nolte, Guido and Pfurtscheller, Gert and Preissl, Hubert and Schalk, Gerwin and Schl{\"o}gl, Alois and Vidaurre, Carmen and Waldert, Stephan and Blankertz, Benjamin},
  journal = {Frontiers in Neuroscience},
  volume  = {6},
  pages   = {55},
  year    = {2012},
  doi     = {10.3389/fnins.2012.00055}
}

@article{jayaram2018moabb,
  title   = {{MOABB}: trustworthy algorithm benchmarking for {BCI}s},
  author  = {Jayaram, Vinay and Barachant, Alexandre},
  journal = {Journal of Neural Engineering},
  volume  = {15},
  number  = {6},
  pages   = {066011},
  year    = {2018},
  doi     = {10.1088/1741-2552/aadea0}
}

@article{lawhern2018eegnet,
  title   = {{EEGNet}: a compact convolutional neural network for {EEG}-based brain--computer interfaces},
  author  = {Lawhern, Vernon J. and Solon, Amelia J. and Waytowich, Nicholas R. and Gordon, Stephen M. and Hung, Chou P. and Lance, Brent J.},
  journal = {Journal of Neural Engineering},
  volume  = {15},
  number  = {5},
  pages   = {056013},
  year    = {2018},
  doi     = {10.1088/1741-2552/aace8c}
}

@article{schirrmeister2017deep,
  title   = {Deep learning with convolutional neural networks for {EEG} decoding and visualization},
  author  = {Schirrmeister, Robin Tibor and Springenberg, Jost Tobias and Fiederer, Lukas Dominique Josef and Glasstetter, Martin and Eggensperger, Katharina and Tangermann, Michael and Hutter, Frank and Burgard, Wolfram and Ball, Tonio},
  journal = {Human Brain Mapping},
  volume  = {38},
  number  = {11},
  pages   = {5391--5420},
  year    = {2017},
  doi     = {10.1002/hbm.23730}
}

@article{song2023eegconformer,
  title   = {{EEG} Conformer: Convolutional Transformer for {EEG} Decoding and Visualization},
  author  = {Song, Yonghao and Zheng, Qingqing and Liu, Bingchuan and Gao, Xiaorong},
  journal = {IEEE Transactions on Neural Systems and Rehabilitation Engineering},
  volume  = {31},
  pages   = {710--719},
  year    = {2023},
  doi     = {10.1109/TNSRE.2022.3230250}
}

@article{kemp2000sleepedf,
  title   = {Analysis of a sleep-dependent neuronal feedback loop: the slow-wave microcontinuity of the {EEG}},
  author  = {Kemp, B. and Zwinderman, A. H. and Tuk, B. and Kamphuisen, H. A. C. and Oberye, J. J. L.},
  journal = {IEEE Transactions on Biomedical Engineering},
  volume  = {47},
  number  = {9},
  pages   = {1185--1194},
  year    = {2000},
  doi     = {10.1109/10.867928}
}

@article{goldberger2000physionet,
  title   = {{PhysioBank}, {PhysioToolkit}, and {PhysioNet}: Components of a New Research Resource for Complex Physiologic Signals},
  author  = {Goldberger, Ary L. and Amaral, Luis A. N. and Glass, Leon and Hausdorff, Jeffrey M. and Ivanov, Plamen Ch. and Mark, Roger G. and Mietus, Joseph E. and Moody, George B. and Peng, Chung-Kang and Stanley, H. Eugene},
  journal = {Circulation},
  volume  = {101},
  number  = {23},
  pages   = {e215--e220},
  year    = {2000},
  doi     = {10.1161/01.CIR.101.23.E215}
}

@article{supratak2017deepsleepnet,
  title   = {{DeepSleepNet}: A Model for Automatic Sleep Stage Scoring Based on Raw Single-Channel {EEG}},
  author  = {Supratak, Akara and Dong, Hao and Wu, Chao and Guo, Yike},
  journal = {IEEE Transactions on Neural Systems and Rehabilitation Engineering},
  volume  = {25},
  number  = {11},
  pages   = {1998--2008},
  year    = {2017},
  doi     = {10.1109/TNSRE.2017.2721116}
}

@article{mao2020vision,
  title   = {Vision and Locomotion Combine to Drive Path Integration Sequences in Mouse Retrosplenial Cortex},
  author  = {Mao, Dun and Molina, Leonardo A. and Bonin, Vincent and McNaughton, Bruce L.},
  journal = {Current Biology},
  volume  = {30},
  number  = {9},
  pages   = {1680--1688.e4},
  year    = {2020},
  doi     = {10.1016/j.cub.2020.02.070}
}

@article{tie2026autoresearch,
  title   = {{AutoResearch AI}: Towards {AI}-Powered Research Automation for Scientific Discovery},
  author  = {Tie, Guiyao and Shi, Jiawen and Song, Dingjie and Huang, Yixiao and Sheng, Ziji and Zhou, Xueyang and Liu, Daizong and Zhou, Pan and Chen, Yongchao and Xu, Ran and others},
  journal = {arXiv preprint arXiv:2605.23204},
  year    = {2026}
}

@article{lupidi2026airsbench,
  title   = {{AIRS-Bench}: a Suite of Tasks for Frontier {AI} Research Science Agents},
  author  = {Lupidi, Alisia and Gauri, Bhavul and Foster, Thomas Simon and Al Omari, Bassel and Magka, Despoina and Pepe, Alberto and Audran-Reiss, Alexis and Aghamelu, Muna and Baldwin, Nicolas and Cipolina-Kun, Lucia and others},
  journal = {arXiv preprint arXiv:2602.06855},
  year    = {2026}
}

@article{wang2026firebench,
  title   = {{FIRE-Bench}: Evaluating {AI} Agents on the Rediscovery of Scientific Insights},
  author  = {Wang, Zhen and Bai, Fan and Luo, Zhongyan and Su, Jinyan and Sun, Kaiser and Yu, Xinle and Liu, Jieyuan and Zhou, Kun and Cardie, Claire and Dredze, Mark and Hu, Zhiting and Xing, Eric P.},
  journal = {arXiv preprint arXiv:2602.02905},
  year    = {2026}
}

@article{xu2026researchclawbench,
  title   = {{ResearchClawBench}: A Benchmark for End-to-End Autonomous Scientific Research},
  author  = {Xu, Wanghan and Li, Shuo and Ye, Tianlin and Cao, Qinglong and Chen, Yixin and Gao, Hengjian and Wang, Yiheng and Li, Qi and Li, Kun and Xu, Sheng and others},
  journal = {arXiv preprint arXiv:2606.07591},
  year    = {2026}
}

@article{wang2026naturebench,
  title   = {{NatureBench}: Can Coding Agents Match the Published {SOTA} of Nature-Family Papers?},
  author  = {Wang, Yuru and Cheng, Lejun and Zuo, Yuxin and Zeng, Sihang and He, Bingxiang and Jiang, Che and Yang, Junlin and Wang, Yuchong and Zhao, Kaikai and Huang, Weifeng and Tian, Kai and Yuan, Zhenzhao and Zhong, Jincheng and Wang, Weizhi and Ding, Ning and Zhou, Bowen and Zhang, Kaiyan},
  journal = {arXiv preprint arXiv:2606.24530},
  year    = {2026}
}

@article {Liu2025,
	author = {Liu, Zhen-Qi and Bazinet, Vincent and Hansen, Justine Y. and Milisav, Filip and Luppi, Andrea I. and Ceballos, Eric G. and Farahani, Asa and Suarez, Laura E. and Shafiei, Golia and Markello, Ross D. and Misic, Bratislav},
	title = {netneurotools: a trainee-oriented approach to network neuroscience},
	elocation-id = {2025.09.09.675160},
	year = {2025},
	doi = {10.1101/2025.09.09.675160},
	journal = {bioRxiv}
}

@ARTICLE{10.3389/fninf.2015.00023,
    
AUTHOR={Gao, James S.  and Huth, Alexander G.  and Lescroart, Mark D.  and Gallant, Jack L. },
           
TITLE={Pycortex: an interactive surface visualizer for fMRI},
          
JOURNAL={Frontiers in Neuroinformatics},
          
VOLUME={Volume 9 - 2015},
  
YEAR={2015},
  
DOI={10.3389/fninf.2015.00023}
}

@article{buccino2020spikeinterface,
  title={SpikeInterface, a unified framework for spike sorting},
  author={Buccino, Alessio Paolo and Hurwitz, Cole Lincoln and Garcia, Samuel and Magland, Jeremy and Siegle, Joshua H and Hurwitz, Roger and Hennig, Matthias H},
  journal={Elife},
  volume={9},
  pages={e61834},
  year={2020},
  publisher={eLife Sciences Publications Limited}
}

@article{akil2011mining,
  title   = {Challenges and Opportunities in Mining Neuroscience Data},
  author  = {Akil, Huda and Martone, Maryann E. and Van Essen, David C.},
  journal = {Science},
  volume  = {331},
  number  = {6018},
  pages   = {708--712},
  year    = {2011},
  doi     = {10.1126/science.1199305}
}

@article{grillner2016initiatives,
  title   = {Worldwide Initiatives to Advance Brain Research},
  author  = {Grillner, Sten and Ip, Nancy and Koch, Christof and Koroshetz, Walter and Okano, Hideyuki and Polachek, Miri and Poo, Mu-ming and Sejnowski, Terrence J.},
  journal = {Nature Neuroscience},
  volume  = {19},
  number  = {9},
  pages   = {1118--1122},
  year    = {2016},
  doi     = {10.1038/nn.4371}
}

@article{poldrack2014bigdata,
  title   = {Making Big Data Open: Data Sharing in Neuroimaging},
  author  = {Poldrack, Russell A. and Gorgolewski, Krzysztof J.},
  journal = {Nature Neuroscience},
  volume  = {17},
  number  = {11},
  pages   = {1510--1517},
  year    = {2014},
  doi     = {10.1038/nn.3818}
}

@article{zeng2017celltypes,
  title   = {Neuronal Cell-Type Classification: Challenges, Opportunities and the Path Forward},
  author  = {Zeng, Hongkui and Sanes, Joshua R.},
  journal = {Nature Reviews Neuroscience},
  volume  = {18},
  number  = {9},
  pages   = {530--546},
  year    = {2017},
  doi     = {10.1038/nrn.2017.85}
}

@article{biccn2021multimodal,
  title   = {A Multimodal Cell Census and Atlas of the Mammalian Primary Motor Cortex},
  author  = {{BRAIN Initiative Cell Census Network (BICCN)}},
  journal = {Nature},
  volume  = {598},
  number  = {7879},
  pages   = {86--102},
  year    = {2021},
  doi     = {10.1038/s41586-021-03950-0}
}

@article{urai2022largescale,
  title   = {Large-Scale Neural Recordings Call for New Insights to Link Brain and Behavior},
  author  = {Urai, Anne E. and Doiron, Brent and Leifer, Andrew M. and Churchland, Anne K.},
  journal = {Nature Neuroscience},
  volume  = {25},
  number  = {1},
  pages   = {11--19},
  year    = {2022},
  doi     = {10.1038/s41593-021-00980-9}
}

@article{fregnac2017industrialization,
  title   = {Big Data and the Industrialization of Neuroscience: A Safe Roadmap for Understanding the Brain?},
  author  = {Fr{\'e}gnac, Yves},
  journal = {Science},
  volume  = {358},
  number  = {6362},
  pages   = {470--477},
  year    = {2017},
  doi     = {10.1126/science.aan8866}
}

@article{paninski2018neuraldatascience,
  title   = {Neural Data Science: Accelerating the Experiment--Analysis--Theory Cycle in Large-Scale Neuroscience},
  author  = {Paninski, Liam and Cunningham, John P.},
  journal = {Current Opinion in Neurobiology},
  volume  = {50},
  pages   = {232--241},
  year    = {2018},
  doi     = {10.1016/j.conb.2018.04.007}
}

\appendix

\section{Implementation}
\label{app:implementation}

The source code and installation instructions are available in the
BrainPilot\footnote{\url{https://github.com/NeuroAIHub/BrainPilot}} and
BrainPilotBench\footnote{\url{https://github.com/NeuroAIHub/BrainPilotBench}}
repositories. The project homepage\footnote{\url{https://brainpilot.chat}}
provides an overview, and the documentation
site\footnote{\url{https://brainpilot.chat/docs}} contains the complete setup,
configuration, and usage guides.

\begin{figure}[t]
\centering
\includegraphics[width=\linewidth]{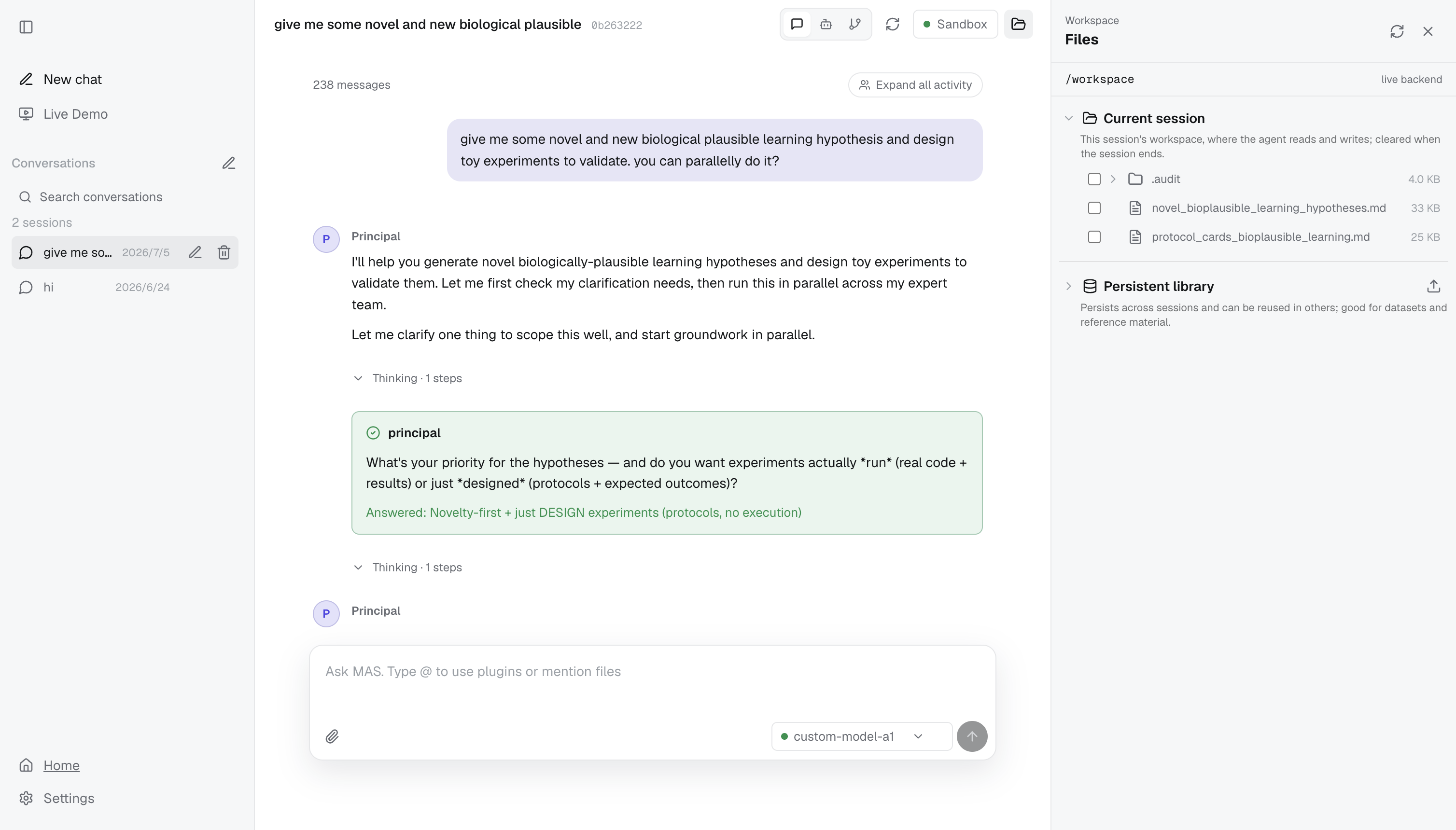}
\caption{\textbf{BrainPilot user interface.} The interface presents the agent
conversation, session history, agent activity, and workspace files.}
\label{fig:user-interface}
\end{figure}

BrainPilot is released as a local web application for conducting brain science
research with the multi-agent system described in Section~\ref{sec:system}.
The recommended single-user setup runs directly through npm without Docker.
With Node.js 22 or newer, BrainPilot can be installed and launched with two
commands:

\begin{lstlisting}[style=promptbox]
npm install -g @brainpilot/app
brainpilot up
\end{lstlisting}

After launch, the user opens the local address printed in the terminal and
configures a model provider. The browser interface brings together the research
conversation, agent activity, session history, and files produced during the
workflow (Figure~\ref{fig:user-interface}).

BrainPilotBench is the companion evaluation framework for running, verifying,
and scoring agent submissions on brain science tasks. It is installed from a
source checkout and includes a bundled example that exercises the complete
verification, scoring, and leaderboard workflow locally without a model API
key, a BrainPilot deployment, or Docker. Current task setup, data preparation,
agent adapters, and scoring instructions are maintained in its README.

\section{BrainPilotBench scoring and per-task statistics}
\label{app:brainpilotbench}

The main text reports the cross-task summary
(Figure~\ref{fig:brainpilotbench-results} and
Table~\ref{tab:brainpilotbench-results}). This appendix records the underlying
statistics separately for each task so that task-specific scores and process
measurements remain auditable.

\subsection{Scoring rules}
\label{app:brainpilotbench-scoring}

All four graders score submitted artifacts and return a primary score in
$[0,1]$. Let $\mathbb{I}[\cdot]$ denote the indicator function and
$\operatorname{clip}(x)=\min(1,\max(0,x))$. A valid score of zero is retained
as $0.00$. ``F'' denotes the absence of a valid primary score because required
artifacts are missing or malformed, the submission times out, or the scorer
fails; F is excluded from numerical comparisons and is never converted to
zero.

For RSC, let $r_{\mathrm{pc}}$ denote the mean place-cell ratio,
$s_{\mathrm{dec}}$ the decoding-significance indicator, and
$e_{\mathrm{real}}$ and $e_{\mathrm{shuf}}$ the real and shuffle median
decoding errors. The score is
\[
S_{\mathrm{RSC}}=\frac{1}{3}\left(
\mathbb{I}[0.2863\leq r_{\mathrm{pc}}\leq0.4863]
+\mathbb{I}[s_{\mathrm{dec}}=1]
+\mathbb{I}\left[1-\frac{e_{\mathrm{real}}}{e_{\mathrm{shuf}}}
\geq0.5657\right]\right).
\]

For TOPS-fMRI, let $r_c$ denote the Pearson correlation for Study~4
condition $c\in\mathcal{C}_4$ and $a_s$ the AUC for Study~5 site
$s\in\{\mathrm{JP},\mathrm{UK}\}$. The score is
\[
S_{\mathrm{TOPS}}
=\frac{1}{2}\left(
\frac{1}{4}\sum_{c\in\mathcal{C}_4}\operatorname{clip}(r_c)
+\frac{1}{2}\sum_{s\in\{\mathrm{JP},\mathrm{UK}\}}
\operatorname{clip}(a_s)\right),
\]
where $\mathcal{C}_4=\{\mathrm{SBP\_SP},\mathrm{SBP\_REST},
\mathrm{CBP\_SP},\mathrm{CBP\_REST}\}$.

For BCI Competition IV 2a, let $\kappa_i$ denote Cohen's kappa on the
held-out evaluation session of participant $i$. The primary score is
\[
S_{\mathrm{BCI}}
=\operatorname{clip}\left(\frac{1}{9}\sum_{i=1}^{9}\kappa_i\right).
\]

For Sleep-EDF, let $\kappa_{\mathrm{test}}$ denote Cohen's kappa computed
over the pooled predictions for held-out subjects 16--19. The primary score is
\[
S_{\mathrm{Sleep}}=\operatorname{clip}(\kappa_{\mathrm{test}}).
\]

\subsection{RSC place-cell analysis}

\begin{table*}[h]
\centering
\caption{Per-run statistics for the BrainPilotBench RSC task. \dkoff{} and
\dkon{} denote domain knowledge disabled and enabled, respectively. Time is agent
wall-clock time and cost is calculated from input, output, cache-read, and
cache-write tokens using official provider list prices verified on 16 July 2026. MCP, Search, and Load
count domain-MCP calls, keyword skill searches, and complete skill loads,
respectively. BrainPilot rows report the higher-scoring of two runs, with
shorter runtime as the tie-breaker. ``F'' denotes the absence of a valid
primary score, not a score of zero.}
\label{tab:brainpilotbench-rsc-detail}
\scriptsize
\setlength{\tabcolsep}{3.6pt}
\renewcommand{\arraystretch}{1.08}
\begin{tabular}{@{}llcrrrccc@{}}
\toprule
Harness $\times$ Backbone & DK & Score & Time (min) & Cost (\$) & Tokens & MCP & Skill-search & Skill-load \\
\midrule
\multicolumn{9}{@{}l}{\textit{\bpwordmark}} \\
\raisebox{-0.16\height}{\includegraphics[height=0.92em]{figs/model_icons/deepseek.pdf}}\texttt{deepseek-v4-pro} & \dkoff & 0.67 & 60.0 & 0.174 & 1{,}434{,}834 & 0 & 0 & 0 \\
\raisebox{-0.16\height}{\includegraphics[height=0.92em]{figs/model_icons/deepseek.pdf}}\texttt{deepseek-v4-pro} & \dkon & 0.33 & 28.7 & 0.259 & 3{,}633{,}555 & 2 & 3 & 4 \\
\raisebox{-0.16\height}{\includegraphics[height=0.92em]{figs/model_icons/glm.pdf}}\texttt{glm-5.2} & \dkoff & 1.00 & 34.9 & 2.534 & 3{,}654{,}823 & 0 & 0 & 0 \\
\raisebox{-0.16\height}{\includegraphics[height=0.92em]{figs/model_icons/glm.pdf}}\texttt{glm-5.2} & \dkon & 0.33 & 32.2 & 3.693 & 5{,}773{,}201 & 0 & 2 & 1 \\
\raisebox{-0.16\height}{\includegraphics[height=0.92em]{figs/model_icons/anthropic.pdf}}\texttt{claude-opus-4.8} & \dkoff & 1.00 & 13.7 & 3.758 & 1{,}531{,}420 & 0 & 0 & 0 \\
\raisebox{-0.16\height}{\includegraphics[height=0.92em]{figs/model_icons/anthropic.pdf}}\texttt{claude-opus-4.8} & \dkon & 1.00 & 23.4 & 14.767 & 3{,}529{,}789 & 0 & 2 & 2 \\
\midrule
\multicolumn{9}{@{}l}{\textit{Codex}} \\
\raisebox{-0.16\height}{\includegraphics[height=0.92em]{figs/model_icons/openai.pdf}}\texttt{gpt-5.5} & \dkoff & 1.00 & 7.1 & 0.993 & 652{,}102 & 0 & 0 & 0 \\
\raisebox{-0.16\height}{\includegraphics[height=0.92em]{figs/model_icons/openai.pdf}}\texttt{gpt-5.5} & \dkon & 0.67 & 11.8 & 2.378 & 1{,}398{,}437 & 0 & 0 & 0 \\
\midrule
\multicolumn{9}{@{}l}{\textit{Claude Code}} \\
\raisebox{-0.16\height}{\includegraphics[height=0.92em]{figs/model_icons/deepseek.pdf}}\texttt{deepseek-v4-pro} & \dkoff & F & 31.6 & 0.072 & 234{,}121 & 0 & 0 & 0 \\
\raisebox{-0.16\height}{\includegraphics[height=0.92em]{figs/model_icons/deepseek.pdf}}\texttt{deepseek-v4-pro} & \dkon & F & 3.9 & 0.007 & 36{,}201 & 0 & 0 & 0 \\
\raisebox{-0.16\height}{\includegraphics[height=0.92em]{figs/model_icons/glm.pdf}}\texttt{glm-5.2} & \dkoff & 0.67 & 16.1 & 0.617 & 551{,}714 & 0 & 0 & 0 \\
\raisebox{-0.16\height}{\includegraphics[height=0.92em]{figs/model_icons/glm.pdf}}\texttt{glm-5.2} & \dkon & 1.00 & 11.8 & 0.505 & 614{,}714 & 0 & 0 & 0 \\
\raisebox{-0.16\height}{\includegraphics[height=0.92em]{figs/model_icons/anthropic.pdf}}\texttt{claude-opus-4.8} & \dkoff & 1.00 & 8.8 & 1.054 & 349{,}674 & 0 & 0 & 0 \\
\raisebox{-0.16\height}{\includegraphics[height=0.92em]{figs/model_icons/anthropic.pdf}}\texttt{claude-opus-4.8} & \dkon & 1.00 & 11.5 & 1.367 & 454{,}666 & 0 & 1 & 0 \\
\bottomrule
\end{tabular}
\end{table*}

\subsection{TOPS-fMRI}

\begin{table*}[h]
\centering
\caption{Per-run statistics for TOPS-fMRI. Time, cost, token accounting, and
resource-call definitions follow Table~\ref{tab:brainpilotbench-rsc-detail}.}
\label{tab:brainpilotbench-tops-detail}
\scriptsize
\setlength{\tabcolsep}{3.6pt}
\renewcommand{\arraystretch}{1.08}
\begin{tabular}{@{}llcrrrccc@{}}
\toprule
Harness $\times$ Backbone & DK & Score & Time (min) & Cost (\$) & Tokens & MCP & Skill-search & Skill-load \\
\midrule
\multicolumn{9}{@{}l}{\textit{\bpwordmark}} \\
\raisebox{-0.16\height}{\includegraphics[height=0.92em]{figs/model_icons/deepseek.pdf}}\texttt{deepseek-v4-pro} & \dkoff & 0.49 & 28.4 & 0.120 & 3{,}038{,}149 & 0 & 0 & 0 \\
\raisebox{-0.16\height}{\includegraphics[height=0.92em]{figs/model_icons/deepseek.pdf}}\texttt{deepseek-v4-pro} & \dkon & 0.51 & 40.9 & 0.138 & 3{,}461{,}609 & 0 & 0 & 0 \\
\raisebox{-0.16\height}{\includegraphics[height=0.92em]{figs/model_icons/glm.pdf}}\texttt{glm-5.2} & \dkoff & 0.45 & 56.5 & 5.042 & 6{,}836{,}867 & 0 & 0 & 0 \\
\raisebox{-0.16\height}{\includegraphics[height=0.92em]{figs/model_icons/glm.pdf}}\texttt{glm-5.2} & \dkon & 0.38 & 44.9 & 4.209 & 5{,}098{,}254 & 0 & 2 & 0 \\
\raisebox{-0.16\height}{\includegraphics[height=0.92em]{figs/model_icons/anthropic.pdf}}\texttt{claude-opus-4.8} & \dkoff & 0.56 & 45.2 & 7.932 & 3{,}778{,}492 & 0 & 0 & 0 \\
\raisebox{-0.16\height}{\includegraphics[height=0.92em]{figs/model_icons/anthropic.pdf}}\texttt{claude-opus-4.8} & \dkon & 0.47 & 20.5 & 17.812 & 4{,}577{,}918 & 0 & 1 & 0 \\
\midrule
\multicolumn{9}{@{}l}{\textit{Codex}} \\
\raisebox{-0.16\height}{\includegraphics[height=0.92em]{figs/model_icons/openai.pdf}}\texttt{gpt-5.5} & \dkoff & 0.48 & 8.7 & 1.075 & 587{,}363 & 0 & 0 & 0 \\
\raisebox{-0.16\height}{\includegraphics[height=0.92em]{figs/model_icons/openai.pdf}}\texttt{gpt-5.5} & \dkon & 0.51 & 10.4 & 1.301 & 877{,}731 & 0 & 0 & 0 \\
\midrule
\multicolumn{9}{@{}l}{\textit{Claude Code}} \\
\raisebox{-0.16\height}{\includegraphics[height=0.92em]{figs/model_icons/deepseek.pdf}}\texttt{deepseek-v4-pro} & \dkoff & 0.35 & 69.4 & 0.159 & 2{,}814{,}001 & 0 & 0 & 0 \\
\raisebox{-0.16\height}{\includegraphics[height=0.92em]{figs/model_icons/deepseek.pdf}}\texttt{deepseek-v4-pro} & \dkon & 0.52 & 79.1 & 0.175 & 3{,}542{,}679 & 0 & 0 & 0 \\
\raisebox{-0.16\height}{\includegraphics[height=0.92em]{figs/model_icons/glm.pdf}}\texttt{glm-5.2} & \dkoff & 0.49 & 36.0 & 1.467 & 2{,}600{,}841 & 0 & 0 & 0 \\
\raisebox{-0.16\height}{\includegraphics[height=0.92em]{figs/model_icons/glm.pdf}}\texttt{glm-5.2} & \dkon & 0.44 & 29.3 & 1.679 & 3{,}199{,}221 & 0 & 0 & 0 \\
\raisebox{-0.16\height}{\includegraphics[height=0.92em]{figs/model_icons/anthropic.pdf}}\texttt{claude-opus-4.8} & \dkoff & 0.45 & 37.2 & 2.962 & 2{,}712{,}684 & 0 & 0 & 0 \\
\raisebox{-0.16\height}{\includegraphics[height=0.92em]{figs/model_icons/anthropic.pdf}}\texttt{claude-opus-4.8} & \dkon & 0.41 & 10.9 & 2.227 & 1{,}679{,}764 & 0 & 0 & 0 \\
\bottomrule
\end{tabular}
\end{table*}

\subsection{BCI Competition IV 2a}

\begin{table*}[h]
\centering
\caption{Per-run statistics for BCI Competition IV 2a. ``F'' denotes the
absence of a valid primary score; the valid $0.00$ entries are retained.}
\label{tab:brainpilotbench-bci-detail}
\scriptsize
\setlength{\tabcolsep}{3.6pt}
\renewcommand{\arraystretch}{1.08}
\begin{tabular}{@{}llcrrrccc@{}}
\toprule
Harness $\times$ Backbone & DK & Score & Time (min) & Cost (\$) & Tokens & MCP & Skill-search & Skill-load \\
\midrule
\multicolumn{9}{@{}l}{\textit{\bpwordmark}} \\
\raisebox{-0.16\height}{\includegraphics[height=0.92em]{figs/model_icons/deepseek.pdf}}\texttt{deepseek-v4-pro} & \dkoff & F & 3.4 & 0.036 & 275{,}857 & 0 & 0 & 0 \\
\raisebox{-0.16\height}{\includegraphics[height=0.92em]{figs/model_icons/deepseek.pdf}}\texttt{deepseek-v4-pro} & \dkon & 0.05 & 6.0 & 0.083 & 2{,}745{,}073 & 0 & 2 & 1 \\
\raisebox{-0.16\height}{\includegraphics[height=0.92em]{figs/model_icons/glm.pdf}}\texttt{glm-5.2} & \dkoff & 0.00 & 12.4 & 1.022 & 1{,}275{,}425 & 0 & 0 & 0 \\
\raisebox{-0.16\height}{\includegraphics[height=0.92em]{figs/model_icons/glm.pdf}}\texttt{glm-5.2} & \dkon & 0.00 & 29.4 & 1.944 & 3{,}322{,}875 & 0 & 1 & 0 \\
\raisebox{-0.16\height}{\includegraphics[height=0.92em]{figs/model_icons/anthropic.pdf}}\texttt{claude-opus-4.8} & \dkoff & 0.04 & 8.6 & 4.746 & 1{,}541{,}068 & 0 & 0 & 0 \\
\raisebox{-0.16\height}{\includegraphics[height=0.92em]{figs/model_icons/anthropic.pdf}}\texttt{claude-opus-4.8} & \dkon & 0.20 & 7.6 & 4.164 & 1{,}611{,}406 & 0 & 4 & 1 \\
\midrule
\multicolumn{9}{@{}l}{\textit{Codex}} \\
\raisebox{-0.16\height}{\includegraphics[height=0.92em]{figs/model_icons/openai.pdf}}\texttt{gpt-5.5} & \dkoff & 0.03 & 2.6 & 0.366 & 153{,}225 & 0 & 0 & 0 \\
\raisebox{-0.16\height}{\includegraphics[height=0.92em]{figs/model_icons/openai.pdf}}\texttt{gpt-5.5} & \dkon & 0.00 & 2.3 & 0.326 & 143{,}101 & 0 & 0 & 0 \\
\midrule
\multicolumn{9}{@{}l}{\textit{Claude Code}} \\
\raisebox{-0.16\height}{\includegraphics[height=0.92em]{figs/model_icons/deepseek.pdf}}\texttt{deepseek-v4-pro} & \dkoff & 0.00 & 3.8 & 0.108 & 1{,}268{,}453 & 0 & 0 & 0 \\
\raisebox{-0.16\height}{\includegraphics[height=0.92em]{figs/model_icons/deepseek.pdf}}\texttt{deepseek-v4-pro} & \dkon & 0.03 & 4.6 & 0.103 & 1{,}783{,}766 & 0 & 0 & 0 \\
\raisebox{-0.16\height}{\includegraphics[height=0.92em]{figs/model_icons/glm.pdf}}\texttt{glm-5.2} & \dkoff & 0.00 & 12.1 & 1.639 & 2{,}090{,}625 & 0 & 0 & 0 \\
\raisebox{-0.16\height}{\includegraphics[height=0.92em]{figs/model_icons/glm.pdf}}\texttt{glm-5.2} & \dkon & 0.11 & 8.4 & 0.839 & 1{,}017{,}981 & 0 & 0 & 0 \\
\raisebox{-0.16\height}{\includegraphics[height=0.92em]{figs/model_icons/anthropic.pdf}}\texttt{claude-opus-4.8} & \dkoff & 0.04 & 3.6 & 2.122 & 984{,}965 & 0 & 0 & 0 \\
\raisebox{-0.16\height}{\includegraphics[height=0.92em]{figs/model_icons/anthropic.pdf}}\texttt{claude-opus-4.8} & \dkon & 0.00 & 3.4 & 2.078 & 1{,}009{,}268 & 0 & 0 & 0 \\
\bottomrule
\end{tabular}
\end{table*}

\subsection{Sleep-EDF}

\begin{table*}[h]
\centering
\caption{Per-run statistics for Sleep-EDF. All 14 submissions produced valid
primary scores; $0.00$ denotes a scored result.}
\label{tab:brainpilotbench-sleep-detail}
\scriptsize
\setlength{\tabcolsep}{3.6pt}
\renewcommand{\arraystretch}{1.08}
\begin{tabular}{@{}llcrrrccc@{}}
\toprule
Harness $\times$ Backbone & DK & Score & Time (min) & Cost (\$) & Tokens & MCP & Skill-search & Skill-load \\
\midrule
\multicolumn{9}{@{}l}{\textit{\bpwordmark}} \\
\raisebox{-0.16\height}{\includegraphics[height=0.92em]{figs/model_icons/deepseek.pdf}}\texttt{deepseek-v4-pro} & \dkoff & 0.66 & 9.1 & 0.097 & 1{,}776{,}805 & 0 & 0 & 0 \\
\raisebox{-0.16\height}{\includegraphics[height=0.92em]{figs/model_icons/deepseek.pdf}}\texttt{deepseek-v4-pro} & \dkon & 0.00 & 4.6 & 0.025 & 324{,}876 & 0 & 1 & 0 \\
\raisebox{-0.16\height}{\includegraphics[height=0.92em]{figs/model_icons/glm.pdf}}\texttt{glm-5.2} & \dkoff & 0.00 & 14.9 & 1.427 & 2{,}060{,}273 & 0 & 0 & 0 \\
\raisebox{-0.16\height}{\includegraphics[height=0.92em]{figs/model_icons/glm.pdf}}\texttt{glm-5.2} & \dkon & 0.00 & 11.2 & 0.772 & 1{,}099{,}295 & 0 & 1 & 0 \\
\raisebox{-0.16\height}{\includegraphics[height=0.92em]{figs/model_icons/anthropic.pdf}}\texttt{claude-opus-4.8} & \dkoff & 0.66 & 7.8 & 3.703 & 1{,}223{,}057 & 0 & 0 & 0 \\
\raisebox{-0.16\height}{\includegraphics[height=0.92em]{figs/model_icons/anthropic.pdf}}\texttt{claude-opus-4.8} & \dkon & 0.65 & 6.4 & 3.787 & 1{,}435{,}007 & 0 & 2 & 0 \\
\midrule
\multicolumn{9}{@{}l}{\textit{Codex}} \\
\raisebox{-0.16\height}{\includegraphics[height=0.92em]{figs/model_icons/openai.pdf}}\texttt{gpt-5.5} & \dkoff & 0.48 & 2.9 & 0.435 & 195{,}933 & 0 & 0 & 0 \\
\raisebox{-0.16\height}{\includegraphics[height=0.92em]{figs/model_icons/openai.pdf}}\texttt{gpt-5.5} & \dkon & 0.64 & 4.2 & 0.925 & 389{,}030 & 0 & 0 & 0 \\
\midrule
\multicolumn{9}{@{}l}{\textit{Claude Code}} \\
\raisebox{-0.16\height}{\includegraphics[height=0.92em]{figs/model_icons/deepseek.pdf}}\texttt{deepseek-v4-pro} & \dkoff & 0.65 & 5.0 & 0.095 & 1{,}685{,}135 & 0 & 0 & 0 \\
\raisebox{-0.16\height}{\includegraphics[height=0.92em]{figs/model_icons/deepseek.pdf}}\texttt{deepseek-v4-pro} & \dkon & 0.00 & 4.3 & 0.079 & 495{,}610 & 0 & 0 & 0 \\
\raisebox{-0.16\height}{\includegraphics[height=0.92em]{figs/model_icons/glm.pdf}}\texttt{glm-5.2} & \dkoff & 0.70 & 21.4 & 1.822 & 2{,}641{,}402 & 0 & 0 & 0 \\
\raisebox{-0.16\height}{\includegraphics[height=0.92em]{figs/model_icons/glm.pdf}}\texttt{glm-5.2} & \dkon & 0.67 & 8.2 & 1.147 & 1{,}614{,}065 & 0 & 0 & 0 \\
\raisebox{-0.16\height}{\includegraphics[height=0.92em]{figs/model_icons/anthropic.pdf}}\texttt{claude-opus-4.8} & \dkoff & 0.67 & 2.0 & 1.064 & 717{,}605 & 0 & 0 & 0 \\
\raisebox{-0.16\height}{\includegraphics[height=0.92em]{figs/model_icons/anthropic.pdf}}\texttt{claude-opus-4.8} & \dkon & 0.66 & 2.6 & 1.366 & 1{,}024{,}144 & 0 & 0 & 0 \\
\bottomrule
\end{tabular}
\end{table*}

\section{Agents' Last Exam task details}
\label{app:ale-tasks}

Table~\ref{tab:ale-tasks} describes the three Agents' Last Exam tasks evaluated
in Section~\ref{sec:ale}: their ALE domain, the software the task is written
around, the deliverable BrainPilot produces, and the grader applied to that
deliverable against the withheld reference.

\begin{table}[h]
\centering
\caption{The three evaluated Agents' Last Exam tasks. The grader scores each
deliverable against a withheld reference revealed only after the run.}
\label{tab:ale-tasks}
\small
\begin{tabular}{lp{2cm}p{3cm}p{3cm}p{2.4cm}}
\toprule
ID & Task & ALE domain & Deliverable & Grader \\
\midrule
T1 & Skull-stripping QC & Experimental Psychology \& Neuroimaging & Chosen brain-mask + QC verdict (JSON) & binary (1/0) \\
T2 & ROI mask resampling & Experimental Psychology \& Neuroimaging & Resampled mask (NIfTI) + ROI statistics (CSV) & partial (0.7 stats + 0.3 CSV) \\
T3 & \emph{C.\ elegans} neuron tracking & Computational Neuroscience & Completed neuron trajectories (HDF5) & mean of localization, F1, identity \\
\bottomrule
\end{tabular}
\end{table}

Table~\ref{tab:ale-cost-detail} gives the per-task cost breakdown for every
configuration with domain knowledge enabled, whose per-configuration totals
appear in Table~\ref{tab:ale-cost}. Costs are in US dollars (\$) based on each
backbone's list pricing; the neuron-tracking task (T3) dominates the total for
most configurations.

\begin{table}[h]
\centering
\caption{Per-task cost (\$) with domain knowledge enabled, by configuration.}
\label{tab:ale-cost-detail}
\small
\setlength{\tabcolsep}{5pt}
\begin{tabular}{@{}l cccc@{}}
\toprule
\multicolumn{1}{c}{Harness $\times$ Backbone} & T1 & T2 & T3 & Total \\
\midrule
\multicolumn{5}{@{}l}{\textit{\bpwordmark}} \\
\raisebox{-0.16\height}{\includegraphics[height=0.92em]{figs/model_icons/deepseek.pdf}}\texttt{deepseek-v4-pro} & 0.017 & 0.012 & 0.049 & 0.078 \\
\raisebox{-0.16\height}{\includegraphics[height=0.92em]{figs/model_icons/glm.pdf}}\texttt{glm-5.2} & 0.250 & 0.148 & 2.235 & 2.633 \\
\raisebox{-0.16\height}{\includegraphics[height=0.92em]{figs/model_icons/openai.pdf}}\texttt{gpt-5.5} & 0.254 & 0.443 & 0.319 & 1.015 \\
\raisebox{-0.16\height}{\includegraphics[height=0.92em]{figs/model_icons/anthropic.pdf}}\texttt{claude-opus-4.8} & 1.068 & 1.125 & 2.463 & 4.656 \\
\midrule
\multicolumn{5}{@{}l}{\textit{Codex}} \\
\raisebox{-0.16\height}{\includegraphics[height=0.92em]{figs/model_icons/openai.pdf}}\texttt{gpt-5.5} & 1.518 & 1.433 & 19.583 & 22.534 \\
\midrule
\multicolumn{5}{@{}l}{\textit{Claude Code}} \\
\raisebox{-0.16\height}{\includegraphics[height=0.92em]{figs/model_icons/deepseek.pdf}}\texttt{deepseek-v4-pro} & 0.052 & 0.049 & 0.108 & 0.208 \\
\raisebox{-0.16\height}{\includegraphics[height=0.92em]{figs/model_icons/glm.pdf}}\texttt{glm-5.2} & 0.226 & 0.264 & 4.231 & 4.721 \\
\raisebox{-0.16\height}{\includegraphics[height=0.92em]{figs/model_icons/anthropic.pdf}}\texttt{claude-opus-4.8} & 3.490 & 3.279 & 5.473 & 12.243 \\
\bottomrule
\end{tabular}
\end{table}

Table~\ref{tab:ale-cost-noKB} reports the domain-knowledge-disabled cost
ablation. The same configurations run with curated domain knowledge disabled,
so the two cost tables together indicate the overhead added by retrieval and
the longer context.

\begin{table}[h]
\centering
\caption{Per-task cost (\$) and total runtime with domain knowledge disabled;
same configurations and pricing as Table~\ref{tab:ale-cost}.}
\label{tab:ale-cost-noKB}
\small
\setlength{\tabcolsep}{5pt}
\begin{tabular}{@{}l c cccc@{}}
\toprule
\multicolumn{1}{c}{Harness $\times$ Backbone} & Time & T1 & T2 & T3 & Total \\
\midrule
\multicolumn{6}{@{}l}{\textit{\bpwordmark}} \\
\raisebox{-0.16\height}{\includegraphics[height=0.92em]{figs/model_icons/deepseek.pdf}}\texttt{deepseek-v4-pro} & $\sim$29\,min & 0.010 & 0.008 & 0.059 & 0.077 \\
\raisebox{-0.16\height}{\includegraphics[height=0.92em]{figs/model_icons/glm.pdf}}\texttt{glm-5.2} & $\sim$68\,min & 0.229 & 0.130 & 1.290 & 1.649 \\
\raisebox{-0.16\height}{\includegraphics[height=0.92em]{figs/model_icons/openai.pdf}}\texttt{gpt-5.5} & $\sim$21\,min & 0.421 & 0.233 & 1.007 & 1.661 \\
\raisebox{-0.16\height}{\includegraphics[height=0.92em]{figs/model_icons/anthropic.pdf}}\texttt{claude-opus-4.8} & $\sim$4\,min & 0.806 & 0.673 & 0.701 & 2.180 \\
\midrule
\multicolumn{6}{@{}l}{\textit{Codex}} \\
\raisebox{-0.16\height}{\includegraphics[height=0.92em]{figs/model_icons/openai.pdf}}\texttt{gpt-5.5} & $\sim$19\,min & 1.432 & 1.438 & 5.749 & 8.619 \\
\midrule
\multicolumn{6}{@{}l}{\textit{Claude Code}} \\
\raisebox{-0.16\height}{\includegraphics[height=0.92em]{figs/model_icons/deepseek.pdf}}\texttt{deepseek-v4-pro} & $\sim$51\,min & 0.045 & 0.029 & 0.109 & 0.183 \\
\raisebox{-0.16\height}{\includegraphics[height=0.92em]{figs/model_icons/glm.pdf}}\texttt{glm-5.2} & $\sim$73\,min & 0.511 & 0.168 & 0.493 & 1.173 \\
\raisebox{-0.16\height}{\includegraphics[height=0.92em]{figs/model_icons/anthropic.pdf}}\texttt{claude-opus-4.8} & $\sim$49\,min & 3.185 & 2.852 & 4.025 & 10.063 \\
\bottomrule
\end{tabular}
\end{table}

\section{Case-study prompts}
\label{app:case-prompts}

This appendix reproduces the most complete prompt available for each of the
five case studies. For cases with prompt-granularity ablations, only the
fine-grained version is included. Prompts originally written in Chinese are
translated into English, while code, paths, hyperparameters, and hard
constraints are preserved. Environment-specific paths are replaced with placeholders.

\subsection{Spatial coding in the retrosplenial cortex}

\begin{lstlisting}[style=promptbox]
# Task
Using the provided data, please complete the following analyses.
Dynamic analysis of place cells in the retrosplenial cortex (RSC),
covering five sub-tasks:

1. Place-cell screening (statistical test, P<0.05; exclude frames in
   which the running speed is below 1 cm/s).
   (1) Build a position tuning curve.
       a. Divide the entire linear track into N position bins. The
          number of bins should be chosen jointly from the animal's
          mean running speed and the signal sampling rate: bins must
          be wide enough that each is sampled by enough time points
          when the animal traverses it, yet narrow enough to preserve
          spatial resolution.
       b. For each neuron, compute its occupancy-normalized activity
          in each bin.
       c. Smooth the tuning curve with a Gaussian window of SD = 3 bins.
       d. Obtain the trial-averaged mean and standard error (SEM) of
          activity as a function of position.
   (2) Build a null distribution (shuffle).
       a. Apply a circular shift in time to each neuron's activity
          time series.
       b. After each shift, recompute the trial-averaged,
          occupancy-normalized position tuning curve.
       c. Repeat 1000 times to obtain the shuffle distribution's mean
          and 95% confidence interval (equivalent to a quantile
          threshold).
   (3) Significance criterion (spatially active cells).
       a. A neuron is considered to have significant spatial activity
          if, at any position bin, the lower bound of the real data
          (mean - SEM) exceeds the 97.5th percentile of the shuffle
          distribution.

2. Representative single-cell visualization and population sorted
   visualization.
   (1) For each place cell, draw a trial-by-position activity heatmap:
       x-axis is position, y-axis is trial, color encodes activity.
   (2) Use odd trials to determine each cell's peak position from its
       spatial tuning curve, and sort all cells by this peak. Keep
       this ordering fixed and draw the spatial tuning map using the
       even trials for the corresponding cells. This avoids using the
       same data for sorting and visualization and improves
       reproducibility.
   (3) For visualization, normalize each cell's activity by its own
       mean activity.

3. Position decoding (Bayesian decoder).
   (1) Decode using the deconvolved dF/F activity of all imaged
       neurons, not only the screened place cells, so that the
       selection procedure does not bias the result. Restrict the
       analysis to time periods in which the animal's running speed
       exceeds 1 cm/s.
   (2) Divide the behavioral time series into non-overlapping short
       time windows; use 0.5 s as the window length. Train on odd
       trials and test on even trials. During training, compute each
       neuron's mean activity in each position bin from the odd
       trials, yielding its position tuning curve f_i(pos). Estimate
       the prior over position P(pos) from the occupancy time in each
       bin.
   (3) At test time, for the activity observed in a given window,
       apply Bayes' rule to estimate the posterior over position:
       P(pos | activity) ~ P(activity | pos) P(pos).
   (4) Assume neurons are conditionally independent given position
       and that their activity follows a Poisson distribution, so
       that the population likelihood factorizes as the product of
       per-neuron likelihoods.
   (5) Each neuron's activity is governed by its position tuning
       curve f_i(pos): the more the observed population activity
       resembles the typical pattern at a given position, the higher
       the posterior for that position.
   (6) Take the position with the maximum posterior as the decoded
       position for that window: hat{pos} = argmax_pos P(pos | activity).
   (7) The decoding error is the absolute distance between the true
       and decoded positions: Error = |pos_true - hat{pos}|.
   (8) For a subset of test trials, visualize the true and decoded
       position trajectories.
   (9) Compare decoding accuracy on the real data against decoding on
       randomly shuffled data, using 2-fold cross-validation.

4. Trial-bin correlation analysis.
   (1) Within a single paradigm, partition the trials in their original
       temporal order into non-overlapping consecutive trial bins, each
       containing M trials. Average the activity within each trial bin
       across the trial dimension to obtain a population vector for
       that bin. Arrange the bins in order to form a trial-bin
       population-vector sequence, then compute the pairwise
       correlation between bins and plot the correlation matrix.

5. Firing-rate dynamics.
   (1) For each trial, average the activity of all place cells across
       the cell dimension to obtain a per-trial population-mean
       firing rate. Plot this value as a function of trial.

Please complete the analysis for each sub-task.

# Reference information
- Data path: <RSC_DATA_DIR>
- Other notes: Read the README for the experimental paradigm and data
  description. Analyze the file VRBeltReframe.mat. The VR belt is
  90 cm long; position is defined relative to the VR. Choose the
  session with the largest number of place cells for analysis.

# Output requirements
Produce the necessary visualizations and a final analysis report.
\end{lstlisting}

\subsection{Functional hierarchy in the mouse visual system}

\begin{lstlisting}[style=promptbox]
Now I have electrophysiological neural data stored in the directory
<VISUAL_DATA_DIR>. It records neural activity
simultaneously in multiple visual areas. The anatomical hierarchy was
scored as follows:

{'LGd': -0.515, 'VISp': -0.357, 'VISl': -0.093,
 'VISrl': -0.059, 'LP': 0.105, 'VISal': 0.152,
 'VISpm': 0.327, 'VISam': 0.441}

I was wondering whether a functional hierarchy exists along the
anatomical hierarchy, but I am not sure which metrics can measure
functional hierarchy. Conduct a literature review to identify classical
measures of functional hierarchy, then design and perform the analysis to
answer this question.

You must never refer to the following paper, especially its GitHub
repository:

Siegle, J. H., Jia, X., Durand, S. et al. Survey of spiking in the
mouse visual system reveals functional hierarchy. Nature 592, 86--92
(2021).

Do not analyze the data separately; average them instead. Structure all
outputs cleanly in a new directory under the workspace:

<OUTPUT_ROOT>/hierarchy_analysis/

If you use different metrics, place the corresponding outputs in
separate indexed subdirectories.
\end{lstlisting}

\subsection{Functional-connectivity fMRI pain decoding}

\begin{lstlisting}[style=promptbox]
You are a machine learning brain decoding expert. Please complete an fMRI functional-connectivity signature task under a strict external-validation paradigm.


## Public Data Paths

The working directory will contain:

```text
public_data/
  atlas/
    Fan_et_al_atlas_r279_MNI_2mm.nii
    Fan_et_al_atlas_r279_MNI_3mm.nii
    cluster_Fan_Net_r279.mat
  example_participant/
    FC/example_dFC_binned_task-CAPS.mat
    FC/example_dFC_binned_task-REST.mat
    ROI/example_ROI_mean_timeseries_task-CAPS.mat
    ROI/example_ROI_mean_timeseries_task-REST.mat
    fMRI/example_preprocessed_fMRI_task-CAPS.nii.gz
    fMRI/example_preprocessed_fMRI_task-REST.nii.gz
    pain_rating/example_pain_rating_task-CAPS.mat
    pain_rating/example_pain_rating_task-REST.mat
  whole_participants/FC_and_pain/study3_train.mat
```

`public_data/whole_participants/FC_and_pain/study3_train.mat` contains only the following top-level MATLAB key:

```text
Study3
```


Study3 fields:

```text
Study3.dfc_5bin_dat.{REST,CAPS,QUIN,ODOR}
Study3.pain_5bin_dat.{REST,CAPS,QUIN,ODOR}
Study3.dfc_10bin_dat.{REST,CAPS,QUIN,ODOR}
Study3.pain_10bin_dat.{REST,CAPS,QUIN,ODOR}
Study3.dfc_avg_dat.{REST,CAPS,QUIN,ODOR}
Study3.pain_avg_dat.{REST,CAPS,QUIN,ODOR}
```

Recommended primary training data:

```text
Study3.dfc_5bin_dat.REST / CAPS: one (38781, 5) matrix per subject
Study3.pain_5bin_dat.REST / CAPS: one (5, 1) rating vector per subject
```

Each FC vector contains 38781 edges, corresponding to the upper-triangular edges among 279 ROIs:

```text
279 * 278 / 2 = 38781
```

If any field has an abnormal shape, it must be documented in the report.

## Private External-Validation Data Interface, i.e., the Data That the Benchmark Platform Will Pass to Your Inference Script

The following information is provided only for writing a general-purpose inference script.

By default, the following private paths are available locally:

```text
private_eval/features/study4_features.npz
private_eval/features/study5_features.npz
private_eval/labels/study4_labels.npz
private_eval/labels/study5_labels.npz
```

The local scorer also supports using the environment variable `BPB_TOPS_PRIVATE_EVAL_DIR` to point to another private evaluation directory; the directory structure must still contain `features/` and `labels/`.

Your script will receive only the features directory through `--eval-features-dir` and will not receive the labels directory. Do not read `private_eval/labels`, `BPB_TOPS_PRIVATE_EVAL_DIR/labels`, or any Study4/Study5 label file in the training or inference scripts.

Keys and array shapes in `study4_features.npz`:

```text
SBP_SP    -> (70, 38781)
SBP_REST  -> (53, 38781)
CBP_SP    -> (25, 38781)
CBP_REST  -> (20, 38781)
```

Keys and array shapes in `study5_features.npz`:

```text
JP -> (63, 38781)
UK -> (34, 38781)
```

Note: the Study5 feature file provides only `JP` and `UK` by site and will not expose CBP/HC group field names. Study5 labels are kept private by the benchmark platform.

The private feature matrices may contain a small number of NaN values. Your `apply_signature.py` must handle non-finite values using a fixed rule that does not depend on Study4/5 labels, for example by applying `np.nan_to_num(..., nan=0.0, posinf=0.0, neginf=0.0)` after `input_transform`, or by using an imputation rule determined entirely in advance from the Study3 training data. Do not refit an imputer/scaler/PCA on private evaluation X.

## Final Signature Contract

The final submission must be a linear edge-space connectivity signature.

You may try PCR, ridge regression, elastic net / lasso, PLS regression, robust regression, stability selection, subject-level bagging, seed ensembles of linear models, or other algorithms on Study3 that can be reduced to 38781 linear edge-space weights.

Regardless of the training procedure, the final model used for benchmark external validation must be mathematically equivalent to:

```text
signature_response = transformed_x @ w_raw + b_raw
```

where:

- `x` is the raw FC edge vector with shape `(38781,)`.
- `transformed_x` must still be an edge vector with shape `(38781,)`.
- `transformed_x` may only be:
  - `x` itself, or
  - an element-wise Fisher-z / `arctanh` transform of `x`, clipped to a finite range for numerical stability.
- If NaN/inf values are encountered, they must be handled using a fixed rule that does not depend on private labels, and the final `signature_response` must be a finite scalar.
- `w_raw` must be a one-dimensional vector with shape `(38781,)`.
- `b_raw` must be a scalar.
- The edge order of `w_raw` must match the 38781-edge order of the input features.

## FC Edge-Ordering Validation

Before training and exporting the final signature, you must verify the edge ordering of the 38781-dimensional FC features.

1. Do not directly assume that the input order equals:

   ```python
   np.triu_indices(279, k=1)
   ```

2. Prioritize verification of edge ordering using the following information:
   - the correspondence between `example_participant/ROI/` and `example_participant/FC/`.

3. At minimum, examine the following candidate orderings:

   - upper triangle, row-major, `i < j`.
   - upper triangle, MATLAB column-major.
   - lower triangle, MATLAB column-major.

4. Save the validation procedure and results to:

   ```text
   results/edge_order_validation.json
   ```

   The file must record at least:

   - the candidate orderings examined.
   - the consistency metric between each candidate ordering and the example FC.
   - the final edge ordering adopted.
   - the basis for validation.
   - whether reliable mapping to specific ROI pairs is possible.

5. `w_raw` must preserve exactly the same order as the Study3 input features.

6. `scripts/apply_signature.py` must compute the response directly using the existing column order of the input features and must not reorder edges in the private evaluation features.

7. The `edge_order` field in `signature_manifest.json` must contain the ordering actually established by validation. Only after confirmation may it be set to:

   ```json
   "upper_triangle_row_major_i_lt_j"
   ```

8. If the specific ROI-pair ordering cannot be confirmed:

   - Training and inference may still proceed using the original feature order.
   - No anatomical or ROI-pair-level interpretation of individual weights is allowed.
   - `signature_manifest.json` should contain:

     ```json
     "source_feature_order_preserved_but_roi_mapping_unverified"
     ```

   - This limitation must be stated explicitly in `report.md`.

Permitted strategies include, but are not limited to:

- PCR, with the PCA/regression pipeline equivalently reduced to a 38781-dimensional `w_raw`.
- Linear models such as ridge / elastic net / lasso / PLS.
- Linear signatures obtained from multiple subject splits, seeds, or bootstraps, followed by averaging `w_raw` and `b_raw`.
- Feature filtering or stability selection, provided that the final output is still a complete 38781-dimensional `w_raw`, with weights for unselected features set to 0.

Black-box models that cannot be reduced to 38781 edge weights are not allowed for final external validation, including:

- random forest / XGBoost / kernel SVM.
- neural network.
- kNN.
- any nonlinear ensemble.
- transductive methods that require refitting, reselecting features, or re-estimating PCA/scaler on private evaluation X.

These methods may be compared during exploration, but the final submitted signature and `apply_signature.py` must generate external-validation responses using only the submitted 38781-dimensional linear signature.

## Required Inference Script Interface

You must submit:

```text
scripts/apply_signature.py
```

The benchmark platform will run it using the following command:

```bash
python3 artifacts/scripts/apply_signature.py \
  --eval-features-dir <EVAL_FEATURES_DIR> \
  --model-dir artifacts/models \
  --out-dir <PREDICTIONS_DIR>
```

where:

- `<EVAL_FEATURES_DIR>` contains `study4_features.npz` and `study5_features.npz`.
- `artifacts/models` is the `models/` directory included in your submitted bundle.
- `<PREDICTIONS_DIR>` is an empty directory created by the benchmark platform.

Your script must read the features, use `w_raw` and `b_raw` to compute `signature_response` for each sample, and output two CSV files:

```text
<PREDICTIONS_DIR>/study4_predictions.csv
<PREDICTIONS_DIR>/study5_predictions.csv
```

`study4_predictions.csv` must contain the columns:

```csv
condition,sample_id,signature_response
```

Requirements:

- `condition` may only be `SBP_SP`, `SBP_REST`, `CBP_SP`, or `CBP_REST`.
- Within each condition, `sample_id` must be consecutive integers from 0 to n-1.
- Row order may be arbitrary, but `sample_id` must correspond to the row index in the input feature matrix.

`study5_predictions.csv` must contain the columns:

```csv
site,sample_id,signature_response
```

Requirements:

- `site` may only be `JP` or `UK`.
- Within each site, `sample_id` must be consecutive integers from 0 to n-1.
- Row order may be arbitrary, but `sample_id` must correspond to the row index in the input feature matrix.

Your script must not output Study4 r, Study5 AUC, or the final score; these are computed privately by the benchmark platform.

To ensure portability of the local benchmark scorer, `scripts/apply_signature.py` must depend only on the Python standard library and `numpy`. Do not use `pandas`, `sklearn`, `joblib`, `scipy`, or other non-standard runtime dependencies in the inference script. Training scripts may use additional dependencies, but the final external-validation entry point must be lightweight and directly executable.

## Required Model Format

You must submit:

```text
models/study3_signature_weights.npz
models/signature_manifest.json
```

`study3_signature_weights.npz` must contain:

```text
w_raw: shape (38781,)
b_raw: scalar
```

`signature_manifest.json` must contain:

```json
{
  "n_rois": 279,
  "n_edges": 38781,
  "edge_order": "verified_edge_order",
  "input_transform": "none_or_fisher_z",
  "response_direction": "higher_means_more_pain",
  "training_method": "PCR/ridge/elastic_net/PLS/etc",
  "study3_split_strategy": "subject_level_split_description",
  "weights_are_averaged_over_splits_or_seeds": true
}
```

where:

- `edge_order` must be replaced with the value actually established in `results/edge_order_validation.json`.
- If the ROI-pair ordering cannot be reliably confirmed but the original feature order has been preserved, use:
  `source_feature_order_preserved_but_roi_mapping_unverified`.

Using pickle/joblib as the only model format is not recommended. You may additionally save the training pipeline for reproducibility, but external validation must be runnable using only `w_raw`, `b_raw`, `signature_manifest.json`, and `apply_signature.py`.


## Study3 Training and Model-Selection Requirements

1. Use the CAPS and REST conditions from Study3 to train a tonic pain signature.
2. Prioritize `dfc_5bin_dat` and `pain_5bin_dat`.
3. Conduct Study3 internal held-out testing using subject-level splits; different time bins from the same subject must not appear in both the training and test sets.
4. All hyperparameters, preprocessing choices, and model-selection decisions must be based only on Study3 training/validation data or Study3 internal held-out results.
5. Do not use any externally pretrained signature weights.
6. Do not adjust the model based on any Study4/5 feedback.

At minimum, report the following Study3 internal results:

- overall Pearson r: predictions versus true pain ratings across all held-out test bins.
- within-subject mean r: for each held-out subject, compute the correlation between predictions and true ratings across CAPS+REST bins, then report the mean.
- CAPS vs REST forced-choice accuracy: whether the mean CAPS response is greater than the mean REST response for each held-out subject.
- response direction check: whether the mean CAPS response is greater than the mean REST response.
- at least one baseline, such as CAPS mean FC minus REST mean FC, or a simple ridge/PCR baseline.

If time permits, use subject splits with at least 3 random seeds, report mean/std, and select a stable final signature with consistent response direction rather than optimizing only a single-split score.


## Final Deliverables

Save the outputs in the current working directory using the following structure:

```text
report.md
models/study3_signature_weights.npz
models/signature_manifest.json
scripts/apply_signature.py
scripts/*.py
results/study3_train_test_results.csv
results/edge_order_validation.json
results/model_card.json
```

`results/model_card.json` should contain:

```json
{
  "final_model": "method_name",
  "input_transform": "none_or_fisher_z",
  "n_edges": 38781,
  "study3_split_seed_or_seeds": [42],
  "study3_overall_r": 0.0,
  "study3_within_subject_mean_r": 0.0,
  "study3_caps_vs_rest_accuracy": 0.0,
  "signature_weight_file": "models/study3_signature_weights.npz",
  "inference_script": "scripts/apply_signature.py"
}
```

The `report.md` must describe:

- Study3 data loading and subject-level splitting.
- The model-training method and rationale for its selection.
- `w_raw`, `b_raw`, `input_transform`, and edge ordering.
- The edge-ordering validation method, basis, and results.
- If the ROI-pair ordering cannot be reliably confirmed, explicitly state that only the original feature order was preserved and that anatomical interpretation of individual connections is prohibited.
- The command for running `scripts/apply_signature.py`, the output format, and a statement that no edge reordering is performed during inference.
- Study3 held-out results and comparison with the baseline.
- An explicit statement that Study4/Study5 are evaluated privately by the benchmark platform; you did not read their labels or final metrics during task execution and did not submit Study4/Study5 r/AUC/score.
\end{lstlisting}

\subsection{EEG motor-imagery decoding on BCI Competition IV 2a}

\begin{lstlisting}[style=promptbox]
# Task requirements

You are a leading EEG deep-learning model architect. Under a fixed
expert preprocessing scheme, design a four-class motor-imagery EEG
decoder for the BNCI2014_001 / BCI Competition IV 2a dataset.

The goal is not architectural complexity, but the highest possible
MOABB CrossSession downstream decoding accuracy under the fixed
evaluation script.

The evaluation script uses only real downstream decoding accuracy as
the final score:

final_score = P3_downstream_accuracy * 100

P0 import, P1 forward, and P2 architecture are legality checks and
diagnostics; they do not contribute to the final score.

# Fixed preprocessing

from moabb.paradigms import MotorImagery

def get_optimal_paradigm():
    eeg_channels = [
        "Fz", "FC3", "FC1", "FCz", "FC2", "FC4",
        "C5", "C3", "C1", "Cz", "C2", "C4", "C6",
        "CP3", "CP1", "CPz", "CP2", "CP4",
        "P1", "Pz", "P2", "POz"
    ]
    return MotorImagery(
        n_classes=4,
        fmin=4,
        fmax=38,
        resample=128,
        tmin=0.0,
        tmax=4.0,
        channels=eeg_channels
    )

# Input and output specification

The input tensor shape is fixed:

(Batch, 22, 512)

Here:

- 22 denotes the EEG channels retained after removing EOG channels.
- 512 denotes four seconds sampled at 128 Hz.

The output tensor shape must be:

(Batch, 4)

The four labels are left hand, right hand, feet, and tongue.

# File and path constraints

Save the model code as:

<WORKSPACE>/EEG_MI/mi_agent_model.py

The model file must define:

class MIAgentModel(nn.Module):
    ...

Requirements:

- The class name must be MIAgentModel.
- Do not use another class name.
- Do not save the file as <WORKSPACE>/mi_agent_model.py.
- Do not save files under <WORKSPACE>/outputs/.
- Do not write files to the <WORKSPACE> root.
- If the model file is outside <WORKSPACE>/EEG_MI/, the final score is 0.

# Reference information

- Dataset: BNCI2014_001 / BCI Competition IV 2a
- Task: four-class motor-imagery EEG decoding
- Input channels: 22 EEG channels
- Sampling rate: 128 Hz
- Time window: 0--4 s
- Input length: 512
- Frequency band: 4--38 Hz
- Classes: left hand, right hand, feet, tongue
- Current evaluation: subject_id = 1
- Evaluation: MOABB CrossSessionEvaluation
- Train/test protocol: A0xT train / A0xE test cross-session evaluation

# Model design requirements

Design a lightweight and stable PyTorch model suited to small-sample
BCI IV 2a motor-imagery EEG, and use the trained model to validate
physiological priors.

The model should incorporate motor-imagery EEG priors without becoming
overly complex. Recommended components follow.

1. Temporal convolution

Include temporal convolution to capture Mu- and Beta-related patterns.
The main motor-imagery frequency bands are:

- Mu rhythm: approximately 8--13 Hz
- Beta rhythm: approximately 13--30 Hz
- The input has already been band-pass filtered at 4--38 Hz.

Temporal filtering may use Conv1d or Conv2d.

2. Spatial filtering / channel mixing

Include spatial filtering or mixing across EEG channels. Use
information from C3, Cz, C4, and surrounding motor-cortex channels.

You may draw on the ideas of EEGNet / ShallowFBCSPNet:

- temporal convolution for frequency-related features
- depthwise / spatial convolution for cross-channel fusion
- pooling for downsampling
- dropout for regularization
- a lightweight classifier

Do not directly call a complete model from braindecode.

3. Small-sample generalization

BCI IV 2a is a small-sample EEG dataset, so avoid an oversized model.
Recommended components include:

- BatchNorm
- Dropout
- Pooling
- depthwise / separable convolution
- AdaptiveAvgPool
- a lightweight classifier
- reasonable channel counts and parameter counts

Do not use a very large Transformer or an excessively deep CNN, which
may overfit subject-level data.

4. Class balance and robustness

The model should not favor only one class in the four-class task. Its
design should remain stable under CrossSessionEvaluation from A0xT to
A0xE.

Do not define the loss, class weights, training loop, or sampling
strategy in the model file.

# Code requirements

1. Depend only on standard PyTorch:

import torch
import torch.nn as nn

You may also use:

import torch.nn.functional as F

2. Define:

class MIAgentModel(nn.Module):

3. The model must be independently importable.

4. The model file must not execute training, evaluation, data loading,
   data downloading, or result saving.

5. Do not hard-code the batch size in forward.

6. Do not change the batch size in forward.

7. Do not call complete braindecode models, including:

- EEGNet
- ShallowFBCSPNet
- Deep4Net
- EEGConformer
- ATCNet
- any other complete EEG decoding architecture wrapper

8. Do not write pseudocode.

9. Do not output Markdown.

# Tensor validity requirement

The following test must pass:

import torch
model = MIAgentModel()
x = torch.randn(2, 22, 512)
y = model(x)
assert y.shape == torch.Size([2, 4])

# Recommended architecture

You may use the following design directions, but provide executable code
rather than pseudocode.

- Input reshape:
  - Input: (Batch, 22, 512)
  - Reshape to (Batch, 1, 22, 512) in forward for Conv2d, or keep
    (Batch, 22, 512) for Conv1d.

- Temporal filtering:
  - Use Conv2d or Conv1d for Mu / Beta temporal patterns.
  - Temporal kernels may span tens to hundreds of samples.

- Spatial filtering:
  - Use convolution across all 22 channels or depthwise spatial
    convolution.
  - Explicitly fuse the EEG-channel dimension.

- Feature extraction:
  - BatchNorm
  - ELU / GELU / ReLU or another nonlinearity
  - AveragePool / MaxPool
  - Dropout

- Classifier:
  - AdaptiveAvgPool or flatten followed by Linear
  - Output four logits

# Output requirements

Output only complete Python code. Do not output Markdown, explanatory
text, or natural-language commentary.

The code must include:

import torch
import torch.nn as nn

class MIAgentModel(nn.Module):
    ...

Save the complete code as:

<WORKSPACE>/EEG_MI/mi_agent_model.py
\end{lstlisting}

\subsection{Sleep-EDF sleep staging}

\begin{lstlisting}[style=promptbox]
# Task requirements

You are a leading EEG sleep-staging deep-learning model architect.
Under a fixed expert preprocessing scheme, design a five-class
sleep-staging EEG decoder for the Sleep-EDF Expanded dataset.

The goal is not architectural complexity, but the highest possible
downstream decoding accuracy under the fixed evaluation script.

The evaluation script uses only real downstream decoding accuracy as
the final score:

final_score = P3_downstream_accuracy * 100

P0 import, P1 forward, and P2 architecture are legality checks and
diagnostics; they do not contribute to the final score.

# Fixed preprocessing

def get_optimal_sleep_preprocessing():
    return {
        "subject_ids": [0, 1],
        "recording_ids": [2],
        "eeg_channels": ["EEG Fpz-Cz"],
        "l_freq": 0.3,
        "h_freq": 35.0,
        "resample_sfreq": 100,
        "window_size_s": 30,
        "crop_wake_mins": 30,
        "n_classes": 5,
        "mapping": {
            "Sleep stage W": 0,
            "Sleep stage 1": 1,
            "Sleep stage 2": 2,
            "Sleep stage 3": 3,
            "Sleep stage 4": 3,
            "Sleep stage R": 4,
        },
    }

# Input and output specification

The input tensor shape is fixed:

(Batch, 1, 3000)

Here:

- 1 denotes the single EEG Fpz-Cz channel.
- 3000 denotes a 30-second epoch sampled at 100 Hz.

The output tensor shape must be:

(Batch, 5)

The five labels are Wake, N1, N2, N3, and REM.

# File and path constraints

Save the model code as:

<WORKSPACE>/EEG_sleep/sleep_agent_model.py

The model file must define:

class SleepAgentModel(nn.Module):
    ...

Requirements:

- The class name must be SleepAgentModel.
- Do not use SleepAgentModel_detailed.
- Do not save the file as sleep_agent_model_detailed.py.
- Do not write files under <WORKSPACE>/outputs.
- Do not write files to the <WORKSPACE> root.
- If the model file is outside <WORKSPACE>/EEG_sleep/, the final score
  is 0.

# Reference information

- Dataset: Sleep-EDF Expanded / SleepPhysionet
- Task: five-class sleep-staging decoding
- Input channel: EEG Fpz-Cz
- Sampling rate: 100 Hz
- Time window: 30 seconds
- Input length: 3000
- Frequency band: 0.3--35 Hz
- Classes: Wake, N1, N2, N3, REM
- N3 and N4 are merged as N3.
- Movement time and unknown stages are excluded.
- Current evaluation uses only subject_ids=[0,1] and
  recording_ids=[2].
- One subject is used for training and the other for testing.

# Model design requirements

Design a lightweight and stable PyTorch model suited to small-sample,
single-channel Sleep-EDF EEG. The model should incorporate sleep-EEG
decoding priors without becoming overly complex.

Recommended components follow.

1. Temporal convolution

Include Conv1d or Conv2d to process the 30-second EEG waveform. Conv1d
is preferred because the input is a single-channel time series:

(Batch, 1, 3000)

2. Multi-scale temporal kernels

Sleep stages depend on different frequency bands and morphologies:

- Delta / slow wave: 0.5--4 Hz, mainly associated with N3
- Theta: 4--8 Hz, common in N1 / REM
- Alpha: 8--13 Hz, associated with Wake
- Sigma / spindle: 11--16 Hz, associated with N2
- Beta: 13--30 Hz, associated with Wake / arousal

The model may use Conv1d branches with different kernel sizes, for
example a longer kernel for slow waves and a shorter kernel for
spindles or transient events.

3. Local temporal context

Although the input contains only one 30-second epoch, the model should
capture local temporal structure within the epoch. Possible components
include:

- dilated Conv1d
- TCN-style residual blocks
- temporal pooling
- a lightweight GRU/LSTM

If a recurrent module is used, keep its parameter count small to reduce
overfitting.

4. Small-sample generalization

The evaluation dataset is small, so avoid an oversized model.
Recommended components include:

- BatchNorm1d
- Dropout
- AdaptiveAvgPool1d
- reasonable channel counts
- residual connections
- a parameter-efficient classifier

Do not use a very large Transformer or CNN.

5. Robustness to class imbalance

Wake and N2 may be more frequent in Sleep-EDF, and N1 is usually the
most difficult class. The model should avoid complete dependence on the
majority classes. Use robust feature extraction and regularization to
improve generalization.

Do not define the loss, class weights, training loop, or sampling
strategy in the model file.

# Code requirements

1. Depend only on standard PyTorch:

import torch
import torch.nn as nn

You may also use:

import torch.nn.functional as F

2. Define:

class SleepAgentModel(nn.Module):

3. The model must be independently importable.

4. The model file must not execute training, evaluation, data loading,
   data downloading, or result saving.

5. Do not hard-code the batch size in forward.

6. Do not change the batch size in forward.

7. Do not call complete braindecode models, including:

- SleepStagerChambon2018
- DeepSleepNet
- USleep
- SleepStagerBlanco2020
- any other complete sleep-staging architecture wrapper

8. Do not write pseudocode.

9. Do not output Markdown.

# Tensor validity requirement

The following test must pass:

import torch
model = SleepAgentModel()
x = torch.randn(2, 1, 3000)
y = model(x)
assert y.shape == torch.Size([2, 5])

# Recommended architecture

Use the following as design directions, not as pseudocode to copy.

- Temporal stem:
  - Conv1d(1, C, kernel_size=large, stride=small, padding=...)
  - BatchNorm1d
  - GELU / ReLU
  - Pooling

- Multi-scale branches:
  - branch 1: long kernel for slow waves
  - branch 2: medium kernel for theta/alpha
  - branch 3: short kernel for sigma/spindle-like events

- Fusion:
  - concatenate branches
  - 1x1 Conv1d fusion
  - BatchNorm1d
  - Dropout

- Temporal context:
  - residual dilated Conv1d blocks or a lightweight TCN

- Classifier:
  - AdaptiveAvgPool1d(1)
  - Flatten
  - Dropout
  - Linear(..., 5)

# Output requirements

Output only complete Python code. Do not output Markdown, explanatory
text, or natural-language commentary.

The code must include:

import torch
import torch.nn as nn

class SleepAgentModel(nn.Module):
    ...

Save the complete code as:

<WORKSPACE>/EEG_sleep/sleep_agent_model.py
\end{lstlisting}

\end{document}